\documentclass[11pt]{article}
\usepackage[preprint]{acl}
\usepackage{times}
\usepackage{latexsym}
\usepackage[T1]{fontenc}
\usepackage[utf8]{inputenc}
\usepackage{multicol}
\usepackage{booktabs} 
\usepackage{longtable} 
\usepackage{rotating}
\usepackage{siunitx}    
\usepackage{amsmath}
\usepackage{algorithm}
\usepackage{algpseudocode}
\usepackage{amsfonts} 
\usepackage{amsthm}
\usepackage{threeparttable}
\usepackage[dvipsnames]{xcolor}   
\definecolor{Teal}{RGB}{0,128,128}
\usepackage{tcolorbox}         
\usepackage{enumitem}
\usepackage{lipsum}    
\usepackage{caption}
\usepackage{multirow} 
\tcbuselibrary{skins}
\usepackage{fontawesome5}
\usepackage[dvipsnames]{xcolor}
\usepackage{enumitem}   
\usepackage{tikz}
\usetikzlibrary{positioning, arrows.meta, backgrounds, shadows}
\usepackage{siunitx}      
\usepackage{booktabs}    
\usepackage{colortbl}     
\usepackage{pifont}
\usepackage{amssymb}
\definecolor{PlausibleGreen}{HTML}{28A745} 
\definecolor{NotPlausibleRed}{HTML}{DC3545} 
\definecolor{DeepSkyBlue4}{RGB}{0, 104, 139}
\newcommand{\cmark}{\textcolor{PlausibleGreen}{\faCheckCircle}}
\newcommand{\xmark}{\textcolor{NotPlausibleRed}{\faTimesCircle}}
\usepackage{microtype}
\usepackage{inconsolata}
\usepackage{amsmath}
\usepackage{graphicx}
\usepackage{wrapfig} 
\usepackage{hyperref}
\hypersetup{% cite style
hidelinks,
colorlinks=true,
urlcolor=DeepSkyBlue4, 
citecolor=cyan
}
\newcommand{\augurIcon}{%
  \tikz[baseline=-0.5ex, scale=0.18]{
    \def\r{1.15}
    \def\ri{0.58}
    \foreach \a/\col in {%
      0/red,30/OrangeRed,60/Orange,90/Goldenrod,%
      120/Yellow,150/LimeGreen,180/SeaGreen,210/Teal,%
      240/Cyan,270/RoyalBlue,300/BlueViolet,330/Magenta% 
    }{%
      \fill[fill=\col, draw=none] (0,0) -- (\a:\r) arc (\a:\a+30:\r) -- cycle;
    }
    \foreach \k in {0,30,...,330}{
      \draw[white, line width=0.28mm] (0,0) -- (\k:\r);
    }
    \draw[white, line width=0.28mm] (0,0) circle (\ri);
    \draw[white, line width=0.28mm] (0,0) circle (\r);
    \foreach \k in {0,60,...,300}{
      \draw[black, line width=0.20mm] (0,0) -- (\k:\ri);
    }
    \fill[white] (0,0) circle (0.14);
  }%
}
\title{~\texorpdfstring{\augurIcon}{Augur}Augur: Modeling Covariate Causal Associations in Time Series via Large Language Models}

\author{
  Zhiqing Cui\textsuperscript{1,3}, \
  Binwu Wang\textsuperscript{1,2}\thanks{Corresponding Authors.}, \
  Qingxiang Liu\textsuperscript{4}, \
  Yeqiang Wang\textsuperscript{5},
  Zhengyang Zhou\textsuperscript{1,2}, \
  \\
  \textbf{Yuxuan Liang\textsuperscript{4}}, \
  \textbf{Yang Wang\textsuperscript{1,2}\footnotemark[1]}
  \\
  \textsuperscript{1}Suzhou Institute for Advanced Research, USTC, Suzhou, China \\
  \textsuperscript{2}University of Science and Technology of China (USTC), Hefei, China \\
  \textsuperscript{3}Nanjing University of Information Science \& Technology, Nanjing, China\\
  \textsuperscript{4}The Hong Kong University of Science and Technology (Guangzhou), Guangzhou, China \\
  \textsuperscript{5}Shanghai Jiao Tong University, Shanghai, China \\
  \texttt{\{zhiqing@nuist.edu.cn, \{wbw2024, angyan\}@ustc.edu.cn\}} \\
  \\
  \\
}

\begin{document}

\maketitle

\begin{abstract}
Large language models (LLM) have emerged as a promising avenue for time series forecasting, offering the potential to integrate multimodal data. However, existing LLM-based approaches face notable limitations—such as marginalized role in model architectures, reliance on coarse statistical text prompts, and lack of interpretability. In this work, we introduce Augur, a fully LLM driven time series forecasting framework that exploits LLM causal reasoning to discover and use directed causal associations among covariates. Augur uses a two stage teacher student architecture where a powerful teacher LLM infers a directed causal graph from time series using heuristic search together with pairwise causality testing. A lightweight student agent then refines the graph and fine-tune on high-confidence causal associations that are encoded as rich textual prompts to perform forecasting. This design improves predictive accuracy while yielding transparent, traceable reasoning about variable interactions. Extensive experiments on real-world datasets with \textbf{26} baselines demonstrate that Augur achieves competitive performance and robust zero-shot generalization.
\end{abstract}

\section{Introduction}

Time series forecasting serves as a critical task for analyzing complex dynamic systems across various domains \cite{wang2024deep,liang2024foundation,qiu2024tfb}. The objective is to predict future time series values by leveraging historical observations collected from target systems and simultaneously observed auxiliary covariate features \cite{wang2024timexer,chen2025select,wang2024timexer}. In recent years, the emergence of Large Language Models (LLMs) has brought transformative opportunities to time series forecasting \cite{jin2024position, kong2025time, liu2025efficient}, utilizing their powerful representational capabilities to integrate multimodal data such as textual information.

However, current LLM-based methods are hindered by several fundamental limitations: 
\ding{182} \underline{\textbf{Marginalized Role}}. LLMs are typically relegated to a peripheral role, serving merely as auxiliary modules that post-process or refine representations generated by a primary forecasting model—rather than acting as the central reasoning engine.
\ding{183} \underline{\textbf{Text Prompts}}. The prompts provided to LLMs convey only coarse-grained statistical summaries (e.g., global means and variances) without encoding structured knowledge of causal among covariates. This restricts the LLM’s ability to apply its native reasoning capabilities to uncover and model complex dynamic interdependencies in the data. 
\ding{184} \underline{\textbf{Interpretability}}. Existing approaches generally lack transparent, systematic mechanisms to reason about variable interactions or trace how specific covariates influence final predictions. This interpretability deficit critically undermines trust and usability in high-stakes domains such as finance and healthcare \cite{jiang2025explainable}.

\begin{figure}[!t]
    \centering
    \includegraphics[width=0.5\textwidth]{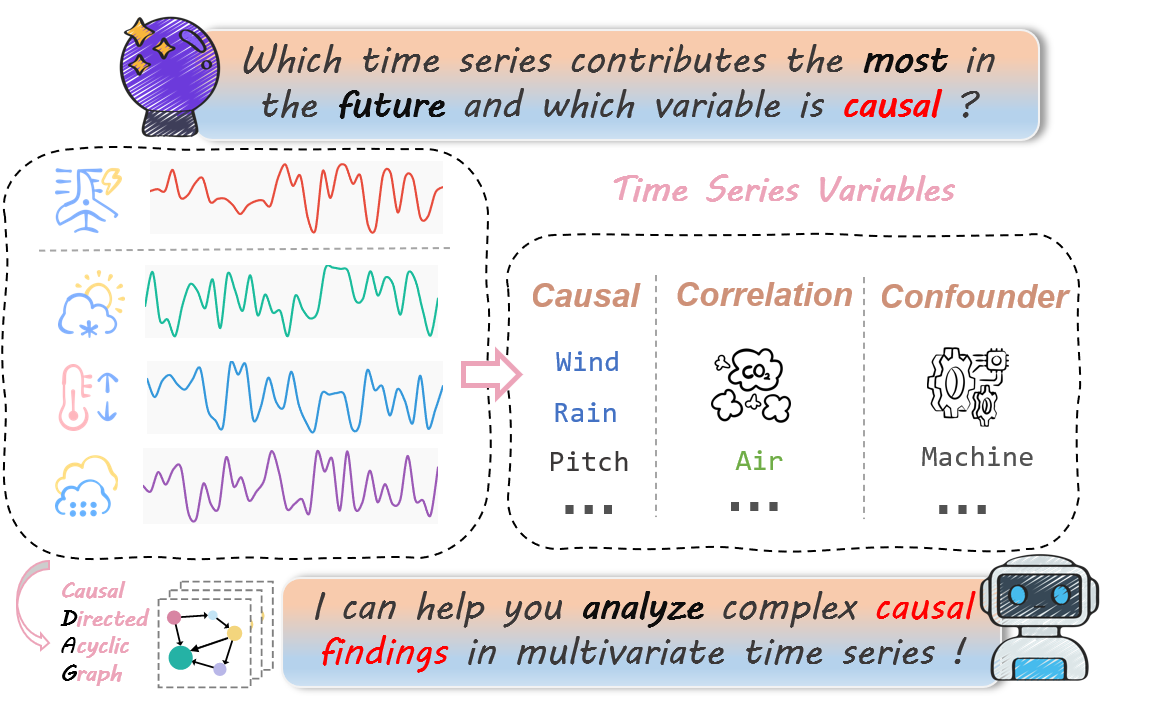}
    \caption{Motivation and problem illustration. Breathing new life into time series tasks with our Augur.}
    \label{fig:graphical_abstract}
    \vspace{-0.15in}
\end{figure}

In this paper, we propose Augur, a novel framework that relies exclusively on LLM for time series forecasting. As illustrated in Figure~\ref{fig:graphical_abstract}, Augur uniquely leverages the causal reasoning capabilities of LLM to uncover latent causal associations among covariates in the time series. This approach not only improves the generalization performance of forecasts but also enhances model interpretability by enabling explicit, traceable reasoning about covariate interactions.

Specifically, Augur employs a two-stage teacher–student architecture. In the first stage, a powerful pre-trained LLM acts as the teacher, identifying potential causal relationships in the time series and encoding them as a directed graph. This process combines a heuristic search-space reduction algorithm with pairwise causality tests, iteratively pruning spurious edges. In the second stage, a lightweight LLM serves as the student, refining the teacher’s graph by retaining only high-confidence causal links. These validated associations, along with their textual summary rather than mere data summaries, are then converted into structured prompts to guide the student’s forecasting. This distillation significantly reduces inference costs and latency compared to deploying the teacher model directly, ensuring practical scalability. Extensive experiments on real-world datasets show that Augur achieves competitive forecasting accuracy and zero-shot generalization.

\vspace{0.1in}

\noindent\textbf{Contribution.} \ding{182} To the best of our knowledge, this work presents the first exploration of LLMs' potential for analyzing causal associations among time series covariates. \ding{183} We propose Augur, a purely LLM-driven time series forecasting framework that leverages a teacher-student dual-stage architecture to refine causal associations and incorporate them as textual prompts, thereby enhancing both predictive accuracy and interpretability. \ding{184} Extensive experiments on real-world datasets with 26 baselines demonstrate that Augur achieves dominant performance.

\begin{figure*}[htbp]
    \centering
    \includegraphics[width=\textwidth]{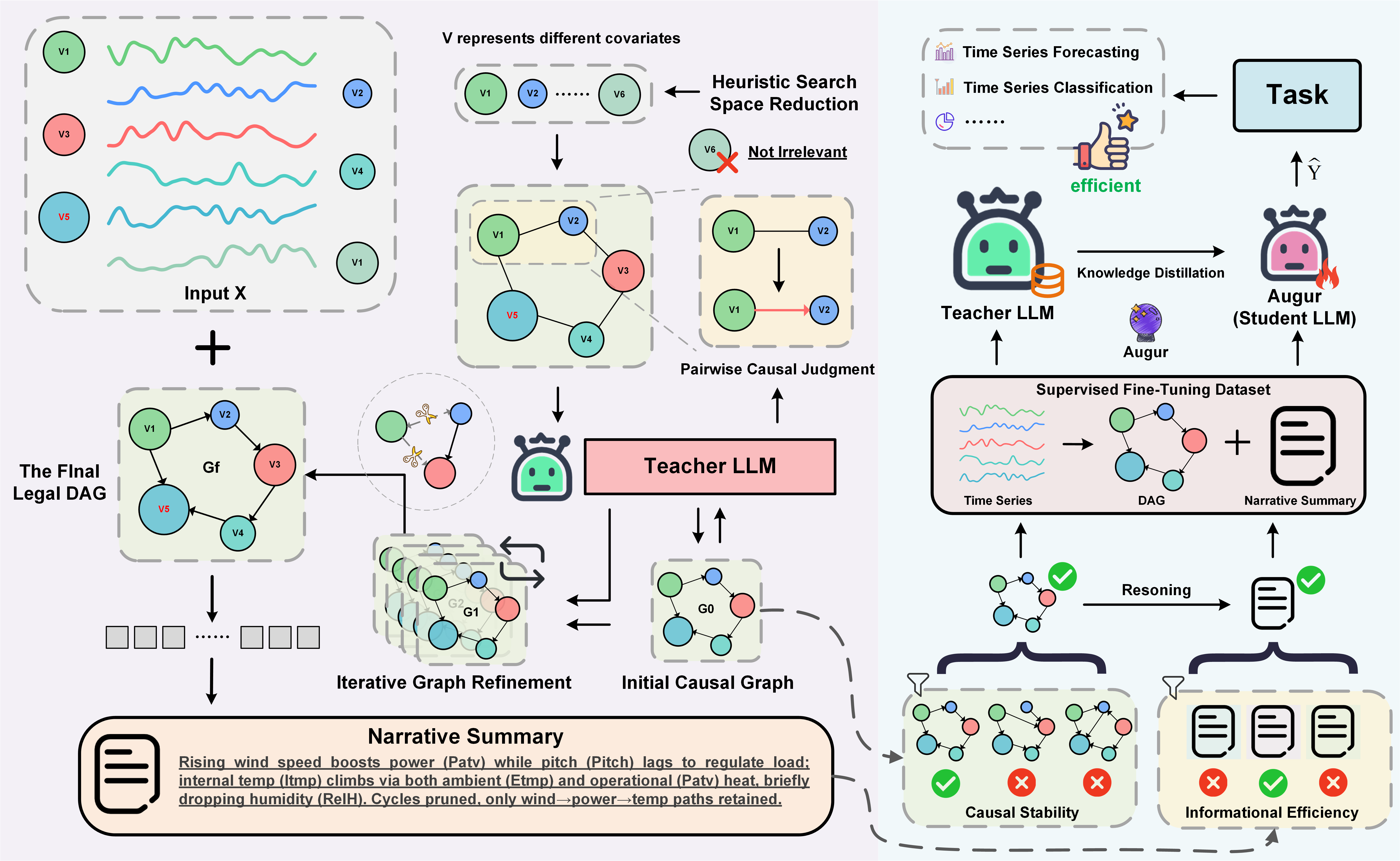}
    \caption{Overview of the Augur framework. Including causal explanation generation and student agent distillation
 for efficient downstream time series tasks.}
    \label{fig:overview}\vspace{-0.1in}
\end{figure*}

\section{Related work}

\paragraph{Time Series Forecasting}
Time series forecasting is a fundamental data analysis task with broad applications across various domains \cite{liang2024foundation,huang2023crossgnn,wang2023pattern,ma2025bist}. Early approaches relied on recurrent models such as Long Short-Term Memory (LSTM) networks and TCN. Recently, Transformer, originally successful in natural language processing and computer vision, is later introduced to time series forecasting \cite{zhou2021informer,zhou2022fedformer,nie2022time}. Furthermore, MLP-based architectures have emerged as lightweight alternatives \cite{zeng2022dlinear,lin2024sparsetsf}. For instance, TimeMixer \cite{wang2024timemixer} achieves competitive performance and remarkable efficiency by combining MLPs with multi-scale modeling. However, these models primarily focus on unimodal temporal dynamics and remain limited in effectively leveraging rich auxiliary modalities such as text.

\paragraph{LLM for Time Series Forecasting}
Recent efforts in time series analysis have increasingly focused on developing general-purpose foundation models, giving rise to two distinct research directions. The first direction aims to build native time series foundation models. This line of work originated with pioneering efforts such as TimeGPT-1~\cite{garza2023timegpt} and has since advanced rapidly, yielding significant innovations—including Chronos’s novel time series tokenization scheme~\cite{ansari2024chronos}, Lag-Llama’s probabilistic forecasting framework~\cite{rasul2023lag}, and massively scaled architectures like TimesFM~\cite{das2024decoder}.

The second direction explores repurposing existing large language models (LLMs) for time series forecasting by bridging the modality gap between numerical sequences and textual representations \cite{tan2024language,gruver2023large}. This stream has evolved from early fine-tuning approaches such as GPT4MTS~\cite{jia2024gpt4mts1111} to more sophisticated, non-invasive alignment strategies—including the reprogramming framework of Time-LLM~\cite{jin2023time} and the instruction-based paradigm of UniTime~\cite{liu2024unitime}—which harness the power of LLMs without modifying their core parameters.

\section{Problem Formulation}
\paragraph{Time Series.} In this work, we focus on the challenge of multimodal time series. Each data instance is represented by a multimodal input pair $(x, s)$, where $x = (x_1, x_2, \dots, x_T) \in \mathbb{R}^{T \times N}$ constitutes the historical sequential observations over a lookback window of length $T$, and $N$ denotes the number of time series covariates. The accompanying component $s$ encapsulates textual data that provides contextual, real-world information pertinent to the numerical observations.

\paragraph{Causal Explanation.} Causal Explanation is denoted as a tuple $(G, S)$, where $G=(V, E)$ is a Causal Directed Acyclic Graph (DAG) used to describe the causal associations between variable pairs, and $E\in \mathbb{R}^{N \times N}$ represents the set of edges. $S$ means Causal Summary that describes the mechanisms encoded in $G$.

\paragraph{Problem Definition} Given historical time series data and its accompanying textual context $(x, s)$, our goal is to predict the future value $y$. This value may represent discrete categories in classification tasks or continuous quantities in regression tasks. Following prior research \cite{zhang2025does,jiang2025explainable}, this study focuses on "discrete trend change" prediction. This emphasis arises because, in decision-critical applications such as risk assessment and strategic planning, understanding the trend direction (e.g., increase, decrease, or drastic change) is often more practically valuable than pursuing precise but uncertain continuous values.

\section{Method}
\label{sec:method}

As shown in Figure \ref{fig:overview} and Algorithm \ref{alg:augur_overall}, our Augur employs a two-stage teacher-student collaborative learning process to produce accurate and interpretable time series predictions.

\noindent\ding{182} \textbf{Causal Explanation Generation via Teacher Model.}
We first utilize a powerful general-purpose LLM foundation model (referred to as the "Teacher" $\mathcal{M}_t$) to perform a preliminary analysis of potential causal associations within massive multivariate time series data. These causal explanations, consisting of a causal graph $G_f$ and a corresponding narrative summary, $S$, along with their associated time series, are distilled into a corpus $\mathcal{D}_{\text{SFT}}$ for supervised fine-tuning the student model. Notably, the teacher model does not undergo any additional fine-tuning stages.

\noindent\ding{183} \textbf{Supervised Fine-tuning of Student Agent.} We begin by refining the corpus generated by the teacher model, eliminating any false or misleading information to ensure the causal explanations are accurate and optimized for downstream tasks. These refined explanations are then utilized for supervised fine-tuning of a smaller, more efficient "Student" agent. This process enables the student model to effectively carry out specific prediction tasks with high accuracy and efficiency.

Based on the above process, a powerful pretrained LLM (e.g., GPT-5) can serve as the teacher model, effectively guiding lightweight student models (e.g., Qwen) designed for specific prediction tasks. This approach fully leverages the strong representation and causal reasoning capabilities of the large teacher model, while significantly reducing deployment and inference costs through the lightweight student model.

\subsection{Causal Explanation Generation via Teacher Model}

\paragraph{Heuristic Search Space Reduction.}
To make the causal discovery process tractable, the teacher model first prunes the combinatorial space of possible edges. Based on the heuristic that significant causal links often produce detectable statistical associations, it computes the Spearman's rank correlation \cite{sedgwick2014spearman} for all variable pairs and forms a candidate set $\mathcal{K}$ of the top-$K$ most correlated pairs:
\begin{equation}
    \mathcal{K} = \text{Top-K}_{(V_a, V_b)} |\rho_s(V_a, V_b)|
\end{equation}
where $\rho_s(\cdot)$ means the Spearman's rank correlation with the time series of node $V_a$ and node $V_b$ as input. This focuses the subsequent, more expensive reasoning process on a high-likelihood subspace of potential causal associations.

\paragraph{Pairwise Causal Judgment.}
Next, the teacher model performs a semantic lift, translating numerical patterns into causal hypotheses. For each candidate pair $(V_a, V_b) \in \mathcal{K}$, the raw time series segments $x_a$ and $x_b$ are serialized into textual representations, $(\tau_a, \tau_b)$, by converting each numerical vector into a comma-separated string. The teacher model then evaluates a discrete hypothesis space $\mathcal{H} = \{V_a \to V_b, V_b \to V_a, \text{Confounded}, \text{Spurious}\}$ to determine the most plausible causal link:
\begin{equation}
    h_{ab}^* = \arg\max_{h_i \in \mathcal{H}} P_{\mathcal{M}_t}(h_i \mid \tau_a, \tau_b)
\end{equation}
The resulting set of directed edges $\{h_{ab}^*\}$ are aggregated to construct an initial global causal graph, $G_0 = (V, E_0)$.

\paragraph{Iterative Causal Graph Refinement.}
The initial graph $G_0$ is treated as a promising but potentially inconsistent hypothesis. The teacher model refines it in an iterative loop to ensure logical consistency. At each step $k$, the agent receives the current graph $G_{k-1}$ and a set of system-generated analytical critiques $C_k$, which are indicators of structural violations (e.g., the presence of a cycle). The teacher then proposes a graph modification $\Delta G_k$ (e.g., an edge reversal or deletion) to resolve the critique:
\begin{equation}
    \Delta G_k = \mathcal{M}_t(G_{k-1}, C_k)
\end{equation}
Specifically, to resolve a cycle, the teacher model is prompted with the full set of edges forming the circular dependency. It then initiates an iterative reasoning process, evaluating the plausibility of each causal link within the context of the entire cycle and its embedded domain knowledge. The model deliberates to identify the link that represents the weakest or least plausible causal associations, designating it for removal. The resulting modification, $\Delta G_k$, represents this reasoned, context-aware decision from the model.
The new graph is formed by $G_k = G_{k-1} \oplus \Delta G_k$, where $\oplus$ denotes the application of the modification to the graph's edge set. This process continues until no critiques remain ($C_k = \emptyset$), yielding a final, validated DAG, $G_f$.

\paragraph{Grounded Narrative Synthesis.}
Finally, the teacher model synthesizes a coherent narrative summary, $S$. It is conditioned on the validated graph $G_f$ and the set of key modifications $\mathcal{I}$ made during refinement (e.g., edges that were removed to break cycles), ensuring the summary is fully grounded in the final causal structure:
\begin{equation}
    S = \mathcal{M}_t(G_f, \mathcal{I})
\end{equation}
We combine the causal explanations generated by the teacher model with their corresponding time series into a corpus, which is denoted as $\mathcal{D}_{\text{SFT}}$. Please note that this corpus only involves a subset of variables from the used datasets.

\begin{algorithm}[t]
\caption{The model process of Augur}
\label{alg:augur_overall}
\begin{algorithmic}[1]
\Require 
    Dataset $\mathcal{D}=\{x_i\}$; 
    Teacher $\mathcal{M}_t$; 
    Student $\mathcal{M}_s^{(0)}$;
    Parameters $K, \lambda, K_{\max}, \tau$
\Ensure Trained student model $\mathcal{M}_s$

\State Initialize SFT dataset: $\mathcal{D}_{\text{SFT}} \gets \emptyset$
\For{each sample $x \in \mathcal{D}$}
    \State $\mathcal{K} \gets \text{Prune}(x, K, \tau)$
    \State $E_0 \gets \text{JudgePairs}(x, \mathcal{K}, \mathcal{M}_t)$
    \State $G_0 \gets (V, E_0)$
    \State $(G_f, \mathcal{I}) \gets \text{Refine}(G_0, \mathcal{M}_t, K_{\max})$
    \State $S \gets \text{Narrate}(\mathcal{M}_t, G_f, \mathcal{I})$
    \State $q \gets \text{Score}(x, G_f, S, \lambda)$
    \If{$q \ge \alpha$}
        \State $\mathcal{D}_{\text{SFT}} \gets \mathcal{D}_{\text{SFT}} \cup \{(x, G_f, S)\}$
    \EndIf
\EndFor
\State $\mathcal{M}_s \gets \text{FineTune}(\mathcal{M}_s^{(0)}, \mathcal{D}_{\text{SFT}})$
\State \Return $\mathcal{M}_s$
\end{algorithmic}
\end{algorithm}

\subsection{Distillation and Training of Student Agent}

After generating a corpus dataset from millions of time series instances, we introduce a distillation process to train a specialized student agent. However, the raw outputs from the teacher model exhibit variable quality and are not uniformly reliable for direct use in training. Therefore, a critical intermediate step is to curate this dataset by scoring and filtering for the highest-quality causal explanations.

\paragraph{Causal Stability ($\mathcal{F}_{\mathrm{s}}$).}
We adopt a consensus-based approach to identify the most robust causal structure. For a given time series $x$, we first generate a set of $N$ diverse candidate DAGs, $\mathcal{G} = \{G_1, G_2, \ldots, G_N\}$ through multiple sampling. We then score each candidate graph $G_k$ based on its structural agreement with all other candidates in the set. The graph with the highest cumulative overlap of causal edges is considered the most stable and reliable explanation. The stability score for a graph $G_k$ with edge set $E_k$ is defined as the sum of shared edges with all other graphs in the ensemble:
\begin{equation}
    \label{eq:reconstruct_stability}
    \mathcal{F}_{\mathrm{s}}(G_k | \mathcal{G}) = \sum_{j=1}^{N} |E_k \cap E_j|
\end{equation}
The final graph selected for the quality function is the one that maximizes this stability score, $G^* = \arg\max_{G_k \in \mathcal{G}} \mathcal{F}_{\mathrm{s}}(G_k | \mathcal{G})$.

\paragraph{Informational Efficiency ($\mathcal{F}_{\mathrm{e}}$).} This term rewards explanations that are both concise and logically grounded. It combines a precision-based Groundedness Score ($S_G$) with a penalty for the summary's length $|S|$. $S_G$ is calculated as the proportion of causal claims extracted from the summary text $S$ that have a corresponding edge in the graph $G$. We employ a lightweight auxiliary model to parse these explicit causal relations. The metric is defined as:
\begin{equation}
    \label{eq:efficiency}
    \mathcal{F}_{\mathrm{e}} = S_G(S, G) -\lambda \cdot |S|
\end{equation}

Finally, we evaluate the overall quality of each causal explanation by considering both causal stability and informational efficiency scores to select only the highest-quality explanations for our training corpus.
\paragraph{Supervised Fine-Tuning.}
Our training data is composed of the curated set of optimal pairs $\{ (x_i, G_i^*, S_i^*) \}_{i=1}^M$. Each target explanation $(G_i^*, S_i^*)$ is serialized into a single text sequence, $Y_i^*$. The student agent is then fine-tuned to map a given time series $x_i$ to its target explanation $Y_i^*$ by minimizing the standard cross-entropy loss:
\begin{equation}
    \label{eq:sft_loss}
    \mathcal{L}_{\text{SFT}} = - \sum_{i} \log P(Y_i^* | x_i; \theta_{\text{s}})
\end{equation}
This distillation process transfers the complex, multi-step reasoning of the teacher into a single, efficient student model. This step is essential to ensure low-latency inference suitable for real-time applications.

\paragraph{Inference Mechanism.}
During the inference phase, the fine-tuned student agent $\mathcal{M}_s$ operates autonomously without further reliance on the teacher. Given a new time series input $x_{new}$, the student directly generates a concise causal rationale abstract based on the learned patterns from $\mathcal{D}_{\text{SFT}}$ and then predicted trend $\hat{y}$. 
\subsection{Utility of the Causal Summary}

The synthesized Causal Summary ($S_g$) provides critical utility by translating the formal, complex Causal DAG ($G_f$) into a human-readable narrative. This validated causal structure then serves as a definitive guide for feature selection, enabling the construction of sparse, robust predictive models based on the true causal drivers (the Markov Blanket) while explicitly excluding known confounders or downstream effects. Furthermore, the summary text $S$ itself becomes a powerful asset; it can be injected back into a multi-modal forecasting model as a rule-based instruction or an informative prior. When provided alongside new numerical data, this text acts as a physics constraint, ensuring the model adheres to the known causal logic to dramatically improve its forecasting accuracy and robustness, especially in novel or out-of-distribution scenarios.

\section{Experiment}
In this section, we conduct extensive experiments to answer the following research questions (RQs):
\vspace{-0.5em}
\begin{itemize}[leftmargin=*]
    \item[\ding{110}] (\textbf{RQ1}) How does our Augur  perform in time series forecasting tasks?  
    \vspace{-0.7em}
    \item[\ding{110}] (\textbf{RQ2}) How is the quality of the causal summaries generated by our Augur?
    \vspace{-0.7em}
    \item[\ding{110}] (\textbf{RQ3}) Is every component of Augur efficient?
    \vspace{-0.7em}
    \item[\ding{110}] (\textbf{RQ4}) Are there marginal effects in causal explanations?
    \vspace{-0.7em}
    \item[\ding{110}] (\textbf{RQ5}) Can our Augur discover physically meaningful causal structures?
\end{itemize}

\subsection{Experiment Setup}
\paragraph{Datasets.}
We employ four time series datasets from diverse real-world domains for evaluation, spanning air, transportation, energy, and finance. These datasets not only contain rich and meaningful covariate features but also exhibit clearly identifiable causal structures. For instance, in the air dataset, meteorological conditions and holiday indicators demonstrate significant and interpretable effects on air pollution dynamics. More details can be found in Appendix \ref{dataset}.
\paragraph{Data Processing.}
We create a hybrid dataset using the LargeAQ air quality dataset \cite{ma2025causalair} and the SDWPF power dataset \cite{zhou2022sdwpf} for causal association analysis in the teacher model and fine-tuning of the student model in our Augur. Our initial training data contains over 100 billion time points and more than 10 million distinct causal events. Traffic and Finance datasets, on the other hand, are directly used to evaluate the model's zero-shot generalization performance.

\paragraph{Task Setting.} Following prior work~\cite{zhang2025does,jiang2025explainable}, we reformulate the forecasting target as a more practical, robust, and interpretable trend prediction task. Specifically, for the four datasets, our task settings are defined as follows. \ding{182} \textbf{Power}: Using the past 24 hours of operational data, we predict whether the wind power output over the next 24 hours will exceed its historical average. \ding{183} \textbf{Air}: Given 48 hours of historical air quality and weather records, we predict whether a severe-level pollution event will occur in the subsequent 24 hours. \ding{184} \textbf{Traffic}: Based on the past 96 hours of data, we classify the average traffic flow trend over the next 24 hours as rising, stable, or falling. \ding{185} \textbf{Finance}: Using the trend from the past four days, we classify whether the stock trend for the next day will be up, down, or neutral. 

\paragraph{Evaluation Metrics.} For prediction tasks, we use two widely used classification metrics: F1-Score and AUROC. To evaluate the quality of the generated causal summaries, we introduced five metrics for comprehensive assessment: BLEU\cite{papineni2002bleu} and ROUGE-L\cite{lin2004rouge} to measure lexical overlap, BERTScore~\cite{zhang2019bertscore} to compare contextual embeddings of the text, Perplexity (PPL) to assess language fluency, and the total length of the summary (in terms of tokens) to evaluate conciseness. Finally, we also conduct human evaluations, with the detailed evaluation criteria provided in Appendix \ref{em2}.

\paragraph{Baselines.} Our experiment compares \textbf{26 advanced baselines}. \ding{182} \textbf{For TS prediction tasks}, we use the latest \underline{\textit{LLM-based models}} such as Time-VLM~\cite{zhong2025time}, Time-LLM~\cite{jin2023time}, GPT4MTS~\cite{jia2024gpt4mts1111}, Moirai~\cite{woo2024unified}, Chronos~\cite{ansari2024chronos}, and Time-MoE~\cite{shi2024time}, as well as \underline{\textit{Classic Unimodal Time Series Models}} including Informer~\cite{zhou2021informer}, Autoformer~\cite{wu2022autoformer}, FEDFormer~\cite{zhou2022fedformer}, DLinear~\cite{zeng2022dlinear}, iTransformer~\cite{liu2024itransformer}, PatchTST~\cite{nie2023patchtst}, LightTS~\cite{campos2023lightts}, TimesNet~\cite{wu2022timesnet}, SparseTSF~\cite{lin2025sparsetsf}, PatchMixer~\cite{gong2023patchmixer}, CycleNet~\cite{lin2024cyclenet}, TimeMixer~\cite{wang2024timemixer}, and TimeXer~\cite{wang2024timexer}. \ding{183} \textbf{For the qualitative evaluation of the causal summaries}, we use powerful LLMs including LLaMA3.1-8B \cite{touvron2023llama}, GPT-4o \cite{hurst2024gpt}, Gemini2.0-flash \cite{team2023gemini}, DeepSeek-v3 \cite{liu2024deepseek},  
Qwen-3-14B \cite{yang2025qwen3}, and ChatTS~\cite{xie2024chatts}.

\paragraph{Implementation.} In the causal explanation extraction stage, we employ \texttt{GPT-5} as the teacher model, utilizing \texttt{gemini-2.5-flash-lite} to execute the initial causal judgment and graph refinement steps. For the fine-tuning phase, \texttt{Qwen3-8b} is adopted as the student agent. The model is optimized using AdamW with a cosine learning rate scheduler, trained for 3 epochs with a global batch size of 64. All datasets undergo strict chronological splitting into training, validation, and test sets following a 7:2:1 ratio. Regarding textual inputs, we utilize a \textit{static} configuration consisting exclusively of dataset descriptions and variable definitions. Crucially, to ensure zero leakage, all models are restricted from accessing any external domain knowledge or future information beyond the historical time series itself. Unimodal baselines rely solely on numerical data. All reported metrics represent the average of five independent runs. Finally, all models were retrained to adapt to these specific tasks. Further implementation details are provided in Appendix~\ref{implement}.

\begin{table*}[!htbp]
\centering
\caption{Performance comparison for Augur on multivariate time series with their pretrained counterparts. Best results are in \textcolor{Magenta}{\textbf{pink}}, and second-best are \textcolor{blue}{\underline{underlined blue}}. }
\label{tab:exp3_zeroshot_expanded_v4}
\resizebox{0.95\textwidth}{!}{%
\begin{tabular}{l cc cc cc cc}
\toprule
& \multicolumn{2}{c}{\textcolor{teal}{\textbf{Air}}} & \multicolumn{2}{c}{\textcolor{orange}{\textbf{Power}}} & \multicolumn{2}{c}{\textcolor{purple}{\textbf{Traffic}}} & \multicolumn{2}{c}{\textcolor{brown}{\textbf{Finance}}} \\
\cmidrule(lr){2-3} \cmidrule(lr){4-5} \cmidrule(lr){6-7} \cmidrule(lr){8-9}
\textbf{Model} & F1-Score & AUROC & F1-Score & AUROC & F1-Score & AUROC & F1-Score & AUROC \\
\midrule
\rowcolor{gray!10}
\quad Time-VLM      & 0.845 & 0.918 & 0.795 & 0.852 & 0.692 & 0.775 & 0.672 & 0.748 \\
\quad Time-LLM     & 0.826 & 0.907 & 0.744 & 0.796 & 0.617 & 0.686 & 0.609 & 0.708 \\
\rowcolor{gray!10}
\quad GPT4MTS       & 0.803 & 0.864 & 0.716 & 0.777 & 0.562 & 0.621 & 0.581 & 0.652 \\
\quad Moirai       & 0.876 & 0.928 & 0.797 & 0.858 & 0.706 & 0.787 & 0.676 & \textcolor{blue}{\underline{0.767}} \\
\rowcolor{gray!10}
\quad Chronos      & 0.839 & 0.921 & 0.789 & 0.849 & 0.698 & 0.769 & 0.665 & 0.744 \\
\quad Time-MoE     & \textcolor{blue}{\underline{0.891}} & \textcolor{blue}{\underline{0.941}} & 0.818 & 0.879 & 0.718 & 0.803 & 0.681 & 0.762 \\
\midrule
\rowcolor{gray!10}
\quad Informer     & 0.846 & 0.903 & 0.764 & 0.844 & 0.676 & 0.748 & 0.646 & 0.717 \\
\quad Autoformer   & 0.803 & 0.908 & 0.757 & 0.836 & 0.684 & 0.756 & 0.653 & 0.724 \\
\rowcolor{gray!10}
\quad FEDFormer    & 0.859 & 0.912 & 0.781 & 0.841 & 0.688 & 0.761 & 0.626 & 0.707 \\
\quad Crossformer  & 0.844 & 0.897 & 0.767 & 0.840 & 0.681 & 0.733 & 0.658 & 0.711 \\
\rowcolor{gray!10}
\quad DLinear      & 0.745 & 0.841 & 0.642 & 0.748 & 0.516 & 0.596 & 0.523 & 0.615 \\
\quad iTransformer & 0.878 & 0.929 & 0.793 & 0.864 & 0.713 & 0.764 & 0.684 & 0.755 \\
\rowcolor{gray!10}
\quad PatchTST     & 0.885 & 0.936 & \textcolor{blue}{\underline{0.823}} & 0.881 & 0.724 & 0.803 & \textcolor{blue}{\underline{0.696}} & 0.748 \\
\quad LightTS      & 0.761 & 0.858 & 0.674 & 0.761 & 0.523 & 0.595 & 0.551 & 0.622 \\
\rowcolor{gray!10}
\quad TimesNet     & 0.864 & 0.919 & 0.759 & 0.851 & 0.703 & 0.751 & 0.667 & 0.736 \\
\quad SparseTSF    & 0.640 & 0.793 & 0.628 & 0.755 & 0.531 & 0.612 & 0.502 & 0.531 \\
\rowcolor{gray!10}
\quad PatchMixer   & 0.805 & 0.869 & 0.702 & 0.768 & 0.542 & 0.613 & 0.568 & 0.659 \\
\quad CycleNet     & 0.811 & 0.874 & 0.708 & 0.773 & 0.538 & 0.609 & 0.524 & 0.635 \\
\rowcolor{gray!10}
\quad TimeMixer    & 0.889 & 0.939 & 0.813 & \textcolor{blue}{\underline{0.884}} & \textcolor{blue}{\underline{0.735}} & \textcolor{blue}{\underline{0.806}} & 0.692 & 0.763 \\
\quad TimeXer      & 0.873 & 0.924 & 0.795 & 0.855 & 0.706 & 0.785 & 0.672 & 0.744 \\
\midrule
\rowcolor{Magenta!10}
\quad \textbf{Augur} & \textcolor{Magenta}{\textbf{0.928}} & \textcolor{Magenta}{\textbf{0.958}} & \textcolor{Magenta}{\textbf{0.849}} & \textcolor{Magenta}{\textbf{0.909}} & \textcolor{Magenta}{\textbf{0.751}} & \textcolor{Magenta}{\textbf{0.825}} & \textcolor{Magenta}{\textbf{0.705}} & \textcolor{Magenta}{\textbf{0.783}} \\
\bottomrule
\end{tabular}%
}
\end{table*}

\subsection{Prediction Performance Study (RQ1)}
We conducted comprehensive experiments to validate the effectiveness of Augur. The experimental results, as shown in Table \ref{tab:exp3_zeroshot_expanded_v4}, demonstrate that Augur achieves the best predictive performance. For traditional time series forecasting models, PatchTST adheres to the principle of independent channel learning, resulting in relatively strong prediction performance. TimeMixer, leveraging a multi-scale modeling approach, effectively captures complex temporal dynamics, thereby achieving the best performance among traditional models while also ensuring competitive zero-shot performance on the Traffic and Finance datasets. Among LLM-based models, Time-MoE achieves the best performance due to its billion-parameter-scale mixture-of-experts architecture, which can effectively capture complex temporal dynamics while maintaining considerable zero-shot generalization capabilities. Our model, Augur, fully leverages the causal inference and analytical capabilities of large models. Its ability to accurately model variable dependencies significantly enhances predictive accuracy. Furthermore, the results on the Traffic and Finance datasets validate its superior generalization performance in zero-shot scenarios.
\begin{table}[h]
\centering
\caption{Human evaluation results: (\textbf{EoU}) ease of understanding, (\textbf{Ins.}) insightfulness for interpretation, and (\textbf{Corr.}) causal correctness.}
\label{tab:human_eval}
\resizebox{0.4\textwidth}{!}{%
\begin{tabular}{@{}llcccc@{}}
\toprule
& & \multicolumn{4}{c}{\textbf{Evaluation Metrics}} \\
\cmidrule(lr){3-6}
\textbf{Dataset} & \textbf{Method} & \textbf{EoU} & \textbf{Ins.} & \textbf{Corr.} & \textbf{Avg.} \\
\midrule
\multirow{3}{*}{\textcolor{orange}{\textbf{Power}}}
 & GPT-4o & 4.3 & 3.9 & \textcolor{blue}{\underline{4.2}} & 4.1 \\
 & Qwen3 & \textcolor{blue}{\underline{4.6}} & \textcolor{blue}{\underline{4.4}} & 3.7 & \textcolor{blue}{\underline{4.2}} \\
 & \textbf{Augur} & \textcolor{Magenta}{\textbf{4.7}} & \textcolor{Magenta}{\textbf{5.2}} & \textcolor{Magenta}{\textbf{4.9}} & \textcolor{Magenta}{\textbf{4.9}} \\
\midrule
\multirow{3}{*}{\textcolor{teal}{\textbf{Air}}}
 & GPT-4o & 4.5 & 4.0 & \textcolor{blue}{\underline{4.3}} & 4.3 \\
 & Qwen3 & \textcolor{Magenta}{\textbf{5.1}} & \textcolor{blue}{\underline{4.5}} & 3.8 & \textcolor{blue}{\underline{4.5}} \\
 & \textbf{Augur} & \textcolor{blue}{\underline{4.9}} & \textcolor{Magenta}{\textbf{5.5}} & \textcolor{Magenta}{\textbf{5.0}} & \textcolor{Magenta}{\textbf{5.1}} \\
\bottomrule
\end{tabular}%
}
\end{table}

\begin{table*}[h]
\centering
\caption{Comprehensive automatic evaluation of summary generation, with datasets on the horizontal axis (\textcolor{orange}{\textbf{Power}} and \textcolor{teal}{\textbf{Air}}). Best results are in \textcolor{Magenta}{\textbf{pink}}, and second-best are \textcolor{blue}{\underline{underlined blue}}.}
\label{tab:exp1_comprehensive_horizontal2}
\resizebox{\textwidth}{!}{%
\begin{tabular}{@{}lccccc|ccccc@{}}
\toprule
& \multicolumn{5}{c|}{\textcolor{orange}{\textbf{Power}}} & \multicolumn{5}{c}{\textcolor{teal}{\textbf{Air}}} \\
\cmidrule(lr){2-6} \cmidrule(lr){7-11}
\textbf{Method} & ROUGE-L & BLEU & BERTScore & PPL & Tokens & ROUGE-L & BLEU & BERTScore & PPL & Tokens \\
\midrule
LLaMA3.1-8B & 0.24 & 0.34 & 0.71 & 34.8 & 1955 & 0.21 & 0.35 & 0.74 & 34.2 & 1751 \\
\rowcolor{gray!10}
GPT-4o & 0.29 & 0.37 & 0.74 & 29.8 & 1854 & 0.29 & 0.38 & 0.79 & 26.5 & 1756 \\
Gemini2.0-Flash & 0.32 & 0.42 & 0.80 & 25.0 & 1365 & 0.36 & 0.47 & 0.82 & 22.5 & 986 \\
\rowcolor{gray!10}
ChatTS-12B & 0.32 & 0.46 & 0.83 & 20.5 & 1134 & 0.33 & 0.43 & 0.84 & 21.2 & 1022 \\
DeepSeek-v3 & 0.34 & \textcolor{blue}{\underline{0.51}} & \textcolor{blue}{\underline{0.87}} & 16.5 & 2010 & 0.37 & \textcolor{blue}{\underline{0.52}} & 0.85 & 15.8 & 1827 \\
\rowcolor{gray!10}
Qwen3-14B & \textcolor{blue}{\underline{0.37}} & 0.45 & 0.84 & \textcolor{blue}{\underline{15.2}} & 1967 & \textcolor{blue}{\underline{0.39}} & 0.48 & \textcolor{blue}{\underline{0.86}} & \textcolor{blue}{\underline{14.7}} & 1788 \\
\midrule
\rowcolor{Magenta!10}
\textbf{Augur} & \textcolor{Magenta}{\textbf{0.49}} & \textcolor{Magenta}{\textbf{0.56}} & \textcolor{Magenta}{\textbf{0.89}} & \textcolor{Magenta}{\textbf{12.3}} & 2317 & \textcolor{Magenta}{\textbf{0.52}} & \textcolor{Magenta}{\textbf{0.57}} & \textcolor{Magenta}{\textbf{0.91}} & \textcolor{Magenta}{\textbf{11.6}} & 2108 \\
\bottomrule
\end{tabular}%
}
\end{table*}

\subsection{Quality of Causcal Summary (RQ2)}
We evaluated the quality of causal summaries generated by various LLMs using the Power and Air datasets. The quantitative metrics, presented in Table \ref{tab:exp1_comprehensive_horizontal2}, and the human evaluation results, detailed in Table \ref{tab:human_eval}, highlight significant differences in performance. Among the models, LLaMA3.1-8B produced the lowest-quality summaries, likely due to its smaller parameter size (8B), which limits its ability to capture complex semantic patterns in multivariate time series data. In contrast, DeepSeek-v3 and Qwen3-14B demonstrated superior causal reasoning capabilities, consistently outperforming other models in summary generation. Our model, Augur, employs a teacher-student two-stage framework to distill and summarize causal associations more effectively. This framework enables Augur to produce higher-quality causal summaries, as validated by both quantitative metrics and human evaluation results. Our model, Augur, leverages a teacher-student two-stage framework to uncover and summarize the underlying causal associations in the data more deeply, thereby generating higher-quality causal summaries.

\subsection{Ablation Study (RQ3)}
In this section, we conduct an ablation study to isolate the contribution of its core components. We create three ablated versions: (1) \textbf{w/o Prune}, which performs causal judgment on all variable pairs without initial filtering; (2) \textbf{w/o Judge}, which bypasses LLM-based causal reasoning and relies on a simpler correlation-based heuristic to orient edges; and (3) \textbf{w/o Refine}, which uses the initial, unrefined graph without ensuring global consistency. 
\begin{table}[!htbp]
\centering
\small
\setlength{\tabcolsep}{3pt}
\caption{Ablation of Augur's core components on the Power dataset.}
\label{tab:ablation_components_with_time}

\sisetup{detect-weight=true, mode=text}

\begin{tabular}{l S[table-format=1.1] ccc S[table-format=1.2] S[table-format=1.2]}
\toprule
\textbf{Variant} & \textbf{Time (×)} & \textbf{Prune} & \textbf{Judge} & \textbf{Refine} & \textbf{F1} & \textbf{AUC} \\
\midrule
w/o Prune  & 5.3 & \xmark & \cmark & \cmark & 0.81 & 0.88 \\
w/o Judge  & 1.7 & \cmark & \xmark & \cmark & 0.76 & 0.84 \\
w/o Refine & 0.6 & \cmark & \cmark & \xmark & 0.79 & 0.87 \\
\midrule
\textbf{Augur} & \bfseries 1.0 & \cmark & \cmark & \cmark & \bfseries 0.85 & \bfseries 0.91 \\
\bottomrule
\end{tabular}
\end{table}
As detailed in Table~\ref{tab:ablation_components_with_time}, removing the initial Prune step significantly increases computational cost, while omitting the LLM-based Judge step leads to the most substantial drop in predictive accuracy. Furthermore, disabling the Refine stage also degrades performance, highlighting the necessity of ensuring a globally coherent causal graph and validating our architectural choices.This design allows us to quantify the necessity of each step—pruning for efficiency, judgment for causal accuracy, and refinement for structural coherence.

\begin{figure*}[!htbp]
    \centering
    \includegraphics[width=\textwidth]{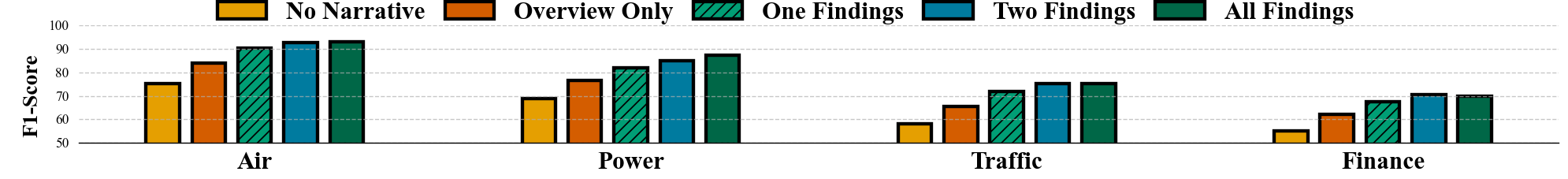}
    \caption{Impact of narrative granularity on zero-shot classification performance across four datasets.}
    \label{fig:narrative_granularity}
\end{figure*}

\begin{figure}[!htbp]
    \centering
    \includegraphics[width=0.7\columnwidth]{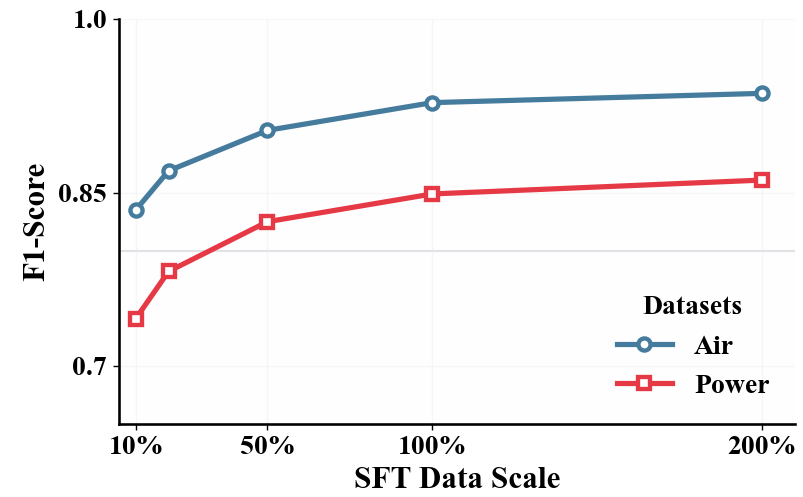}
    \caption{Impact of the SFT data scaling up.}
    \label{fig:sft_data_scale}
\end{figure}

\subsection{Analysis of Narrative Granularity (RQ4)}
We conduct an additional analysis to evaluate the relationship between the granularity of causal explanations and downstream task performance. In our zero-shot forecasting tasks on the Traffic and Finance datasets, we systematically varied the textual inputs, constructing variants ranging from raw time series data with only high-level summaries to progressively more detailed causal discoveries. This approach allowed us to quantify the marginal utility of each additional discovery.

As shown in Figure~\ref{fig:narrative_granularity}, model performance improves significantly when at least one key causal discovery is included, compared to using only high-level summaries. Further analysis reveals a clear point of diminishing returns: while the first two causal discoveries contribute substantial gains, adding a third or subsequent minor discoveries provides negligible benefits.

Figure~\ref{fig:sft_data_scale} indicates that a carefully selected subset of high-quality data yields greater performance improvements than simply expanding the dataset volume. Scaling the supervised fine-tuning corpus to 200\% proves disproportionately costly and, according to our analysis, fails to enhance results. This supports our hypothesis that for this task, a quality-focused "less-is-more" strategy outperforms data-intensive approaches.

\subsection{Causal Discovery Capability (RQ5)}
\label{app:causal_quality}

\noindent\textbf{Physical-based Scenarios.} To validate the causal discovery capability, we strictly evaluated the quality of the causal graphs generated by Augur against a ground truth dataset derived from \textit{representative scenarios} in the \texttt{Power} domain. Since real-world time series lack explicit causal labels, we annotated a subset of variables based on well-established physical principles of wind turbines (e.g., rigid mechanisms like $\mathit{WindSpeed} \to \mathit{ActivePower}$ and thermodynamics like $\mathit{ExternalTemp} \to \mathit{InternalTemp}$). We acknowledge that causal mechanisms in dynamic systems are state-dependent; thus, the specific set of active edges in the ground truth allows for variation or selection contingent upon the specific temporal context of each scenario.

We benchmarked Augur against five classic causal discovery methods: PC \cite{gong2024causal}, GES, PCMCI \cite{runge2022causal2}, Granger Causality \cite{shojaie2022granger}, and Mutual Information (MI) \cite{kraskov2004estimating}. The evaluation metrics include edge-level Precision, Recall, F1-Score, and Structural Hamming Distance (SHD), where a lower SHD indicates a structure closer to the physical ground truth.

\begin{table}[h]
    \centering
    \caption{Quantitative comparison of causal discovery performance on the physics-based ground truth subset.}
    \label{tab:causal_discovery_metrics}
    \small
    \setlength{\tabcolsep}{5pt}
    \begin{tabular}{l cccc}
    \toprule
    \textbf{Method} & \textbf{Precision} & \textbf{Recall} & \textbf{F1-Score} & \textbf{SHD} $\downarrow$ \\
    \midrule
    PC & 0.15 & 0.10 & 0.12 & \textcolor{blue}{\underline{11.8}} \\
    GES & 0.15 & 0.26 & 0.19 & 18.7 \\
    MI & 0.09 & \textcolor{Magenta}{\textbf{0.95}} & 0.17 & 76.1 \\
    Granger & 0.15 & \textcolor{blue}{\underline{0.54}} & \textcolor{blue}{\underline{0.24}} & 26.6 \\
    PCMCI & \textcolor{blue}{\underline{0.16}} & 0.38 & 0.22 & 21.2 \\
    \midrule
    \textbf{Augur} & \textcolor{Magenta}{\textbf{0.41}} & 0.31 & \textcolor{Magenta}{\textbf{0.34}} & \textcolor{Magenta}{\textbf{9.3}} \\
    \bottomrule
    \end{tabular}
\end{table}

As presented in Table~\ref{tab:causal_discovery_metrics}, Augur significantly outperforms traditional statistical methods. While methods like MI achieve high recall by capturing all correlations, they suffer from extremely low precision. In contrast, Augur achieves the lowest SHD and the highest F1-Score , demonstrating its ability to filter spurious correlations and recover cleaner, physically meaningful causal structures.

\vspace{1ex}
\noindent\textbf{Controlled Synthetic.}
Since real-world datasets lack ground truth for every edge, we further conducted a controlled experiment using a noisy Vector Autoregression (VAR) model to generate synthetic data with known causal structures (8 variables, including lagged terms and confounders). This allows for a precise evaluation of edge-level accuracy.

\begin{table}[h]
    \centering
    \caption{Performance on the synthetic dataset with ground-truth causal structures.}
    \label{tab:synthetic_causal_metrics}
    \small
    \setlength{\tabcolsep}{5pt}
    \begin{tabular}{l cccc}
    \toprule
    \textbf{Method} & \textbf{Precision} & \textbf{Recall} & \textbf{F1-Score} & \textbf{SHD} $\downarrow$ \\
    \midrule
    PC & 0.32 & 0.27 & 0.29 & 12.1 \\
    GES & 0.47 & 0.63 & 0.54 & 10.3 \\
    MI & 0.34 & 0.40 & 0.37 & 11.9 \\
    Granger & 0.28 & \textcolor{Magenta}{\textbf{0.92}} & 0.43 & 22.7 \\
    PCMCI & \textcolor{blue}{\underline{0.53}} & \textcolor{blue}{\underline{0.75}} & \textcolor{blue}{\underline{0.62}} & \textcolor{Magenta}{\textbf{5.2}} \\
    \midrule
    \textbf{Augur} & \textcolor{Magenta}{\textbf{0.72}} & 0.58 & \textcolor{Magenta}{\textbf{0.64}} & \textcolor{blue}{\underline{9.4}} \\
    \bottomrule
    \end{tabular}
\end{table}

The results on synthetic data (Table~\ref{tab:synthetic_causal_metrics}) corroborate our findings. While specific baselines like Granger Causality achieve high recall by over-predicting edges, and PCMCI shows strong structural recovery, Augur dominates in Precision and F1-Score. This high precision is particularly critical for our Teacher-Student framework, as it ensures that the student model is fine-tuned on high-confidence, valid causal prompts rather than noisy false positives.

\section{Conclusion}
In this paper, we present Augur, a framework that leverages large language models to extract explicit causal associations among covariates, thereby enhancing both forecasting capability and interpretability. The method implements a teacher-student architecture  to generate high-quality causal explanations. These explanations are subsequently utilized as textual prompts to guide the student LLM in making predictions. By distilling complex reasoning capabilities into a lightweight agent, Augur balances computational efficiency with high-fidelity interpretability, offering a scalable solution for real-world applications that require both accuracy and transparency.

\section*{Limitations}
Our approach fundamentally relies on the assumption of causal sufficiency (i.e., no unobserved confounders) and employs a correlation-based heuristic that may overlook complex non-linear or lagged dependencies. Second, the quality of the generated narratives is contingent on the teacher LLM's internal knowledge, which may be incomplete or biased in highly specialized domains.

\section*{Ethics Statement}
All datasets and language models used in this work are publicly available and comply with relevant licensing terms. No personally identifiable information (PII) or sensitive data was collected or used.Five annotators with formal training in logic provided informed consent and were fairly compensated. Evaluation protocols included clear rubrics to minimize subjective bias. All evaluated data was anonymized.

\bibliography{main}
\appendix
\section*{Appendix} 
This appendix provides supplementary material organized into several sections to support the main paper. Each section is dedicated to a specific topic:

\begin{itemize}
    \item \textbf{Appendix A} establishes the theoretical foundations with a self-contained overview of Causal Directed Acyclic Graphs (DAGs).

    \item \textbf{Appendix B} presents a formal proof of causal feature optimality, building upon the theoretical groundwork.

    \item \textbf{Appendix C} details our comprehensive methodology and implementation, including dataset and baseline descriptions, evaluation protocols, and the technical environment.

    \item \textbf{Appendix D} offers supplementary results and analyses, featuring a quantitative evaluation of our feature selector and an in-depth case study that illustrates the entire pipeline, from initial analysis to the final causal narrative and the evaluation using an LLM-as-a-Judge to validate causal narrative quality.

    \item \textbf{Appendix E} concludes with a discussion on framework extensibility which includes social impact and finally, presents the specific prompts used to guide the LLM-driven processes.
\end{itemize}

\section{Causal Directed Acyclic Graphs}
\label{app:dags}

This appendix provides a brief overview of the key concepts from the theory of causal inference based on Directed Acyclic Graphs (DAGs) that are used in this paper.

\subsection{DAGs and Observational Distributions}
A causal Directed Acyclic Graph (DAG) is a graph $\mathcal{G}=(V,E)$ where nodes $V=\{X_1,\dots,X_n\}$ represent random variables and directed edges $E$ represent direct causal associations, with no directed cycles. The graph structure encodes the \emph{causal Markov property}: every variable is assumed to be independent of its non-descendants given its direct causes (parents), denoted $\mathrm{Pa}_{\mathcal{G}}(X_i)$. This property implies that the joint observational distribution $P(V)$ factorizes according to the graph:
\begin{equation}
P(V) = \prod_{i=1}^n P(X_i \mid \mathrm{Pa}_{\mathcal{G}}(X_i)).
\label{eq:app-markov-factorization}
\end{equation}

\subsection{Interventions and Causal Effects}
A causal effect is defined via a surgical intervention on the system, formalized by Pearl's $do$-operator. An intervention $do(X_k=x)$ sets the variable $X_k$ to a constant value $x$, severing the influence of its natural parents. This corresponds to a modified graph where all edges into $X_k$ are removed. The post-interventional distribution is obtained via a \emph{truncated factorization}:
\begin{equation}
P(V \mid do(X_k)) = {\prod_{i \neq k} P(X_i \mid \mathrm{Pa}_{\mathcal{G}}(X_i)) }.
\label{eq:app-truncated-factorization}
\end{equation}

\subsection{Paths, d-Separation, and Confounding}
Associations between variables in a DAG are transmitted along paths. A \emph{back-door path} from a treatment $X$ to an outcome $Y$ is a path that begins with an edge into $X$ (e.g., $X \leftarrow \dots$). Such paths are non-causal and can create spurious associations due to common causes (\emph{confounders}). A node on a path is a \emph{collider} if both edges on the path point into it (e.g., $A \to C \leftarrow B$). The concept of \emph{d-separation} determines conditional independence: a set of nodes $Z$ d-separates $X$ and $Y$ if it blocks every path between them. A path is blocked by $Z$ if it contains either (1) a non-collider that is in $Z$, or (2) a collider that is not in $Z$ and has no descendants in $Z$. If all paths are blocked, then $X \perp\!\!\!\perp Y \mid Z$.

\subsection{Identifiability via the Back-door Criterion}
The causal effect $P(y \mid do(x))$ can be identified from observational data if confounding can be appropriately controlled. The \emph{back-door criterion} provides a sufficient condition for this. A set of variables $Z$ satisfies the back-door criterion relative to $(X, Y)$ if: (1) no node in $Z$ is a descendant of $X$, and (2) $Z$ blocks every back-door path between $X$ and $Y$. If such a set exists, the causal effect is identifiable via the \emph{back-door adjustment formula}:
\begin{equation}
P(y \mid do(x)) = \sum_{z} P(y \mid x,z)\,P(z).
\label{eq:app-backdoor}
\end{equation}

\subsection{A Consolidated Example}

Consider a causal model represented by the DAG with edges $\{Z \to X, Z \to Y, X \to Y, X \to C \leftarrow Y\}$. Here, $X$ is the treatment and $Y$ is the outcome.
\begin{itemize}
    \item \textbf{Confounding:} The path $X \leftarrow Z \to Y$ is a back-door path created by the common cause (confounder) $Z$. It induces a spurious association between $X$ and $Y$. To estimate the causal effect of $X$ on $Y$, this path must be blocked.
    \item \textbf{Collider:} The node $C$ is a collider. The path $X \to C \leftarrow Y$ is naturally blocked. Conditioning on $C$ would open this path, inducing a spurious association, and is therefore incorrect.
    \item \textbf{Adjustment:} The set $\{Z\}$ satisfies the back-door criterion. $Z$ is not a descendant of $X$, and conditioning on $Z$ blocks the back-door path $X \leftarrow Z \to Y$. Therefore, the causal effect is identifiable by adjusting for $Z$:
    \begin{equation}
      P(y | \text{do}(x)) = \sum_{z} P(y | x, z) P(z).
    \end{equation}
\end{itemize}

\section{A Proof of Causal Feature Optimality}
\label{proof}
We provide a rigorous, first-principles proof that the mutual information between a set of features and a target variable is maximized when the feature set is the target's causal Markov Blanket.

\newtheorem{theorem}{Theorem}
\begin{theorem}
\end{theorem}
Given a system of variables $V$ and a target $Y$ with a known causal DAG $\mathcal{G}$, let $X_c = \mathrm{MB}_{\mathcal{G}}(Y)$ be the causal Markov Blanket of $Y$. Then, the mutual information between $X_c$ and $Y$ is equal to the mutual information between the entire system $V$ and $Y$:
\begin{equation}
I(X_c; Y) = I(V; Y)
\end{equation}

\begin{proof}
The proof relies on showing the equality of the conditional entropies, $H(Y \mid V) = H(Y \mid X_c)$, from which the theorem follows directly from the definition of mutual information, $I(A; B) = H(B) - H(B \mid A)$.

The conditional entropy $H(Y \mid V)$ is defined as:
\begin{multline}
\label{eq:h_y_v_def}
H(Y \mid V) = \\
- \sum_{v \in V} p(v) \sum_{y \in Y} p(y \mid v) \log p(y \mid v)
\end{multline}
By the causal Markov property, $Y$ is conditionally independent of all variables in $V \setminus X_c$ given $X_c$. This implies that for any realization $v$ of $V$, where $v_c$ is the portion corresponding to $X_c$:
\begin{equation}
\label{eq:ci_property_prob}
p(y \mid v) = p(y \mid v_c)
\end{equation}
Substituting \eqref{eq:ci_property_prob} into \eqref{eq:h_y_v_def}, we obtain:
\begin{multline}
\label{eq:h_y_v_substituted}
H(Y \mid V) = \\
- \sum_{v \in V} p(v) \sum_{y \in Y} p(y \mid v_c) \log p(y \mid v_c)
\end{multline}
We can now regroup the summation over all $v \in V$ by summing over the components $v_c \in X_c$ and $v' \in V \setminus X_c$, and then apply the law of total probability to marginalize over $v'$:

\begin{align}
\label{eq:marginalization_steps}
H(Y \mid V) &= - \sum_{v_c, v'} p(v_c, v') \cdot \nonumber \\
& \qquad \left( \sum_{y} p(y|v_c) \log p(y|v_c) \right) \nonumber \\
&= - \sum_{v_c} \left( \sum_{v'} p(v_c, v') \right) \cdot \nonumber \\
& \qquad \left( \sum_{y} p(y|v_c) \log p(y|v_c) \right) \nonumber \\
&= - \sum_{v_c} p(v_c) \left(\sum_{y} p(y|v_c) \log p(y|v_c) \right)
\end{align}
The final expression in \eqref{eq:marginalization_steps} is precisely the definition of the conditional entropy $H(Y \mid X_c)$. Thus, we have shown:
\begin{equation}
\label{eq:entropy_equality_final}
H(Y \mid V) = H(Y \mid X_c)
\end{equation}
From this equality, it directly follows that:
\begin{equation}
\label{eq:mi_equality_final}
H(Y) - H(Y \mid V) = H(Y) - H(Y \mid X_c)
\end{equation}
which proves the theorem that $I(V; Y) = I(X_c; Y)$.
\end{proof}

\section{Experimental Setup}
\label{Experimental Setup}
\begin{figure}[htbp]
    \centering
    \includegraphics[width=0.5\textwidth]{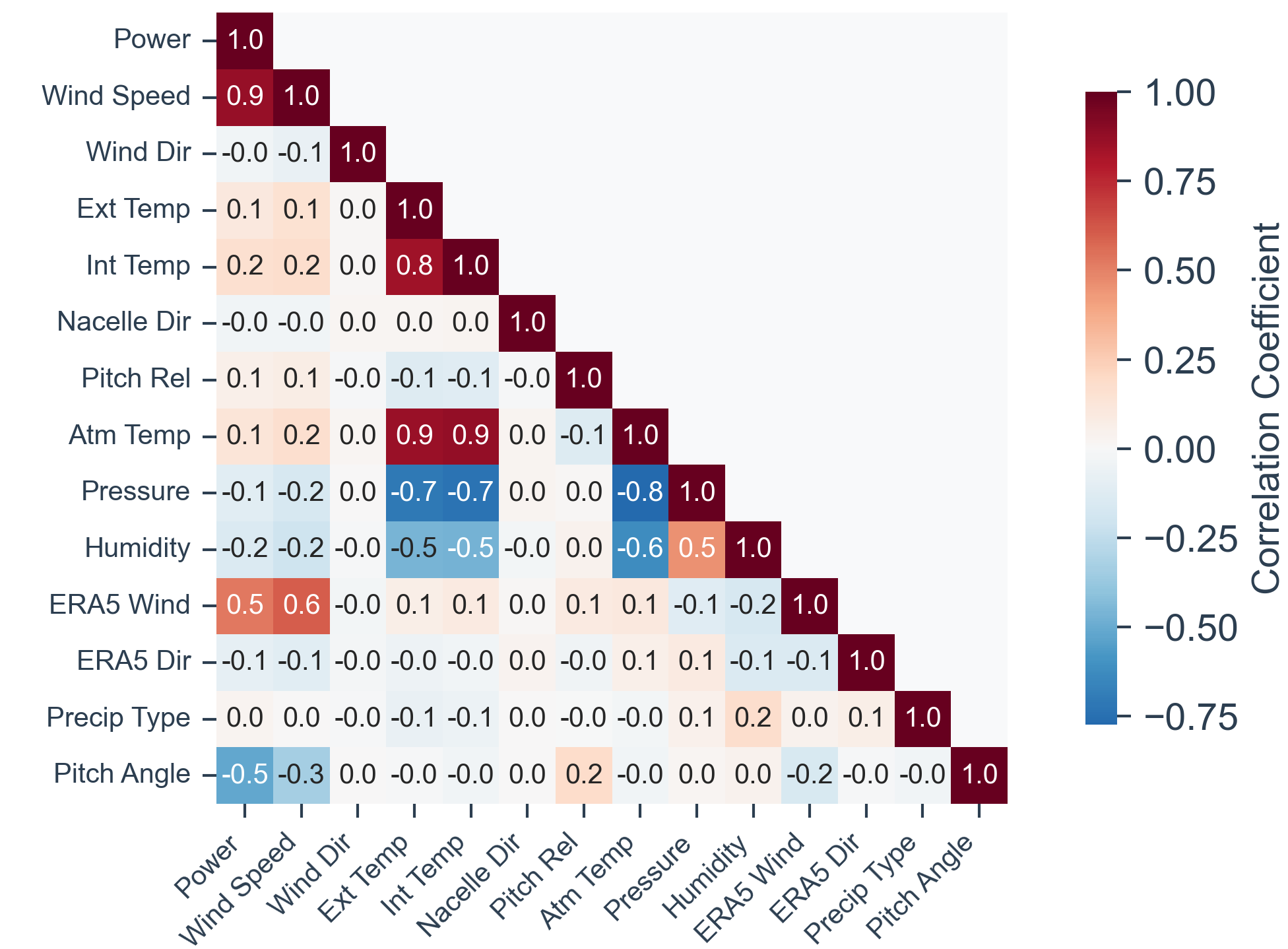}
    \caption{Spearman correlation matrix of the variables in the power dataset.}
    \label{fig:correlation_matrix}
\end{figure}

\subsection{Description of Datasets}\label{dataset}

The \textbf{Power} dataset is sourced from \textbf{SDWPF}~\cite{zhou2022sdwpf}, which was collected over two years (2020–2021) from a wind farm comprising 134 turbines. It contains over 11 million high-resolution records (sampled every 10 minutes), integrating SCADA sensor measurements with ERA5 meteorological reanalysis data. To visualize inter-variable relationships, we compute the Spearman correlation matrix, shown in Figure~\ref{fig:correlation_matrix}. This matrix reveals dependency patterns among several key covariates, providing valuable guidance for our causal discovery process.

The \textbf{Air} dataset is from \textbf{LargeAQ}, a  nationwide air-quality dataset spanning eight years (2015--2023). Each station provides time-stamped observations of major \emph{criteria pollutants} together with rich \emph{meteorological covariates}. Records are provided at (predominantly) hourly cadence, enabling long-horizon AQI research and spatiotemporal modeling.

The \textbf{Traffic} dataset is from \textbf{NZ-Traffic} \cite{li2024high}, which spans a nine-year period. It aggregates data from 2,042 sensors across New Zealand's highway network, encompassing over 600 million high-resolution records (15-min intervals). Each entry provides granular vehicle counts, distinguishing between light-duty and heavy-duty vehicles. This core traffic data is fused with rich contextual information, including key meteorological covariates (e.g., temperature, precipitation) from NOAA and extensive metadata detailing highway structure, coastlines, and public holidays.

The \textbf{finance} dataset is from \textbf{FNSPID} \cite{dong2024fnspid}, spans nearly a quarter-century (1999–2023). It covers 4,775 companies from the SP 500 index and comprises over 29.7 million stock price records alongside 15.7 million financial news articles sourced from four major outlets.

 In our experiments, we use the most important variable as targets and the other variables as covariates; station metadata are used only for grouping and reporting and are not injected as numeric inputs unless explicitly noted.
 
\subsubsection{Evaluation Metrics} \label{em2}

\paragraph{Automatic Evaluation.}
Our evaluation of summary quality is comprehensive. First, to measure content similarity, we compare our generated summaries against reference texts using a suite of metrics. We employ the classic n-gram-based metrics, BLEU~\cite{papineni2002bleu} and ROUGE-L~\cite{lin2004rouge}, to assess lexical overlap. To capture deeper semantic meaning and properly handle paraphrasing, we also include BERTScore~\cite{zhang2019bertscore}, which compares the contextual embeddings of the texts. Second, we assess linguistic fluency using Perplexity (PPL). A lower perplexity score indicates that the generated text is more coherent, grammatically sound, and aligns well with the patterns of natural language.
Finally, we evaluate conciseness by measuring the summary's total length in tokens.

\paragraph{Human Evaluation.}
We conduct a human study to assess how readers perceive causal summaries produced by different methods. For each dataset, we randomly sample 50 instances per method. Five annotators with formal training in logic (at least undergraduate level) independently rate each summary on a 7-point scale along three dimensions: (1) ease of understanding, (2) insightfulness for interpreting the time series, and (3) causal correctness in reflecting inter-variable relationships. 

\subsection{Description of Baselines}
\label{sec:baselines_ref }

Unless otherwise specified, all baseline implementations, data pipelines, and default hyperparameters follow the open-source library \textbf{MM-TSFlib}\cite{liu2024timemmd}\footnote{\url{https://github.com/AdityaLab/MM-TSFlib}}. For Time-LLM and similar models, we follow the authors’ public implementations. We keep their official training/evaluation protocols for both numeric-only time-series models and LLM-aligned variants to ensure a fair and reproducible comparison.

\paragraph{Process.}
We adopt the unified preprocessing: time alignment to a single grid, forward-fill then mean-impute missing values, per-variable z-score standardization with training-split statistics with strictly non-overlapping temporal splits to avoid leakage.

\paragraph{Optimization.}
Our supervised fine-tuning (SFT) is implemented using the LLaMA-Factory framework. We used the AdamW optimizer with a cosine learning rate scheduler. The initial learning rate was set to $2 \times 10^{-5}$, with a warmup ratio of 0.1 and a weight decay of 0.01. We fine-tuned the model for 3 epochs with a global batch size of 64.

For time series forecast model, we use AdamW with cosine decay and $5\%$ warmup, gradient clipping at $1.0$, mixed precision, and early stopping on validation loss (patience $10$). Learning rate is selected from $\{1\mathrm{e}{-3},\,5\mathrm{e}{-4},\,1\mathrm{e}{-4}\}$ for numeric-only baselines and $\{2\mathrm{e}{-4},\,1\mathrm{e}{-4},\,5\mathrm{e}{-5}\}$; batch size from $\{32,\,64,\,128\}$ subject to memory. Max epochs $100$ with model selection on validation performance. Probabilistic baselines use $100$ samples to form point estimates.
\label{implement}
\paragraph{Other models.} 

For the causal \emph{judge} and \emph{refine} stages, each iteration’s decisions (edge proposals and cycle-resolution edits) are generated by a lightweight language model, which we use to produce pairwise causal labels and graph updates until convergence. Since the reasoning process is decomposed into fine-grained, low-complexity atomic tasks, the difficulty is minimal; thus, this component can be effectively substituted with any efficient language model.

We query the vendors’ official APIs. For open-source models, we run \texttt{Qwen-3-8B}, \texttt{ChatTS-14B-0801}\footnote{\url{https://github.com/NetManAIOps/ChatTS}} and \texttt{LLaMA-3.1-8B} locally.
Unless otherwise stated, the OpenAI endpoint uses \texttt{gpt-4o-2024-08-06} (GPT-4o). We apply the same decoding configuration to \texttt{Gemini-2.5-Flash-Lite} and \texttt{gpt-5-mini},We set \texttt{max\_tokens}$=4096$ and use \texttt{temperature}$=0.5$ for analysis-style generation (self-reflection and textual refinement), and \texttt{temperature}$=0.1$ for prediction and causal judgments, which yielded the most stable empirical behavior in our preliminary tests. All prompts share identical instruction templates across providers, with only minimal schema-specific tokens adjusted for compatibility. All evaluations use greedy decoding, and we retain all other settings from the official HuggingFace configurations.

\paragraph{Cost.}
The construction of our SFT corpus involves querying the commercial teacher model (\texttt{GPT-5}) at an average cost of \${1}--\${2} per instance. Crucially, this represents a one-time fixed cost for training data synthesis. In contrast, the deployment of the fine-tuned student model offers significant economic advantages. Our comparative analysis indicates that using the "Full Teacher" for direct inference would incur a recurring cost of approximately \${0.62} per iteration. By distilling this capability into the student agent, we reduce the marginal inference cost to \textbf{$<\${0.01}$}, representing a reduction of over 50$\times$. Furthermore, the student model reduces inference latency by approximately 10$\times$, making Augur a scalable solution for high-frequency forecasting scenarios where direct reliance on frontier LLMs would be financially and computationally prohibitive. We commit to releasing the full synthetic corpus under an open license to facilitate future research.

\subsection{Human Evaluation}
In our human evaluation, we compare GPT-4o, Qwen3, and Augur on the  power and air datasets. For each dataset and system, we uniformly sample 50 instances. Three annotators—each with formal training in logic and basic knowledge of meteorology or energy systems—independently rate every summary. For each annotator, the item order is independently randomized. Ratings follow a 7-point Likert scale across three dimensions: Ease of Understanding (EoU), Insightfulness (Ins.), and Causal Correctness (Corr.).

\subsection{Environment}
All experiments were conducted on a TensorEX server equipped with two Intel Xeon Gold 5218R CPUs, and four NVIDIA A100 80GB GPUs.

\subsection{Rationale for Supervised Fine-Tuning}
Our supervised fine-tuning (SFT) dataset consists of input-target pairs serialized into single text sequences. This format trains the model to map numerical time series and a prompt to a structured causal explanation.

The input sequence contains the numerical data and task instruction, delineated by the \texttt{<|data|>} and \texttt{<|task|>} tokens. The target sequence contains the ground-truth causal graph and a summary, separated by the \texttt{<|graph|>} and \texttt{<|summary|>} tokens. The \texttt{<|EOT|>} token marks the end of both sequences.
\section{Supplementary Results and Analyses}
\label{analysi}
\subsection{Additional Experiments}

\subsubsection{Effectiveness of Causal Feature Selection}
\begin{table*}[htbp]
\centering
\caption{Comparison of Forecasting Performance. The best result in each column is in \textbf{bold}.}
\label{tab:forecasting_results_final}
\begin{tabular}{l rr rr rr rr}
\toprule
& \multicolumn{2}{c}{\textbf{MLP}} & \multicolumn{2}{c}{\textbf{LSTM}} & \multicolumn{2}{c}{\textbf{DLinear}} & \multicolumn{2}{c}{\textbf{PatchTST}} \\
\cmidrule(lr){2-3} \cmidrule(lr){4-5} \cmidrule(lr){6-7} \cmidrule(lr){8-9}
\textbf{Feature Set} & \multicolumn{1}{c}{MSE} & \multicolumn{1}{c}{MAE} & \multicolumn{1}{c}{MSE} & \multicolumn{1}{c}{MAE} & \multicolumn{1}{c}{MSE} & \multicolumn{1}{c}{MAE} & \multicolumn{1}{c}{MSE} & \multicolumn{1}{c}{MAE} \\
\midrule
All Features      & 0.2416          & 0.3770          & 0.1124          & 0.1743          & 0.1591          & 0.2898          & \textbf{0.1144} & \textbf{0.1804} \\
Wind Speed        & 0.2140          & 0.3245          & 0.1505          & 0.2614          & 0.1579          & 0.2782          & 0.1590          & 0.2799 \\
\midrule
\textbf{Augur}    & \textbf{0.2015} & \textbf{0.3110} & \textbf{0.1108} & \textbf{0.1725} & \textbf{0.1560} & \textbf{0.2755} & 0.1252          & 0.1915 \\
\bottomrule
\end{tabular}
\end{table*}

\begin{figure}[htbp]
    \centering
    \includegraphics[width=0.48\textwidth]{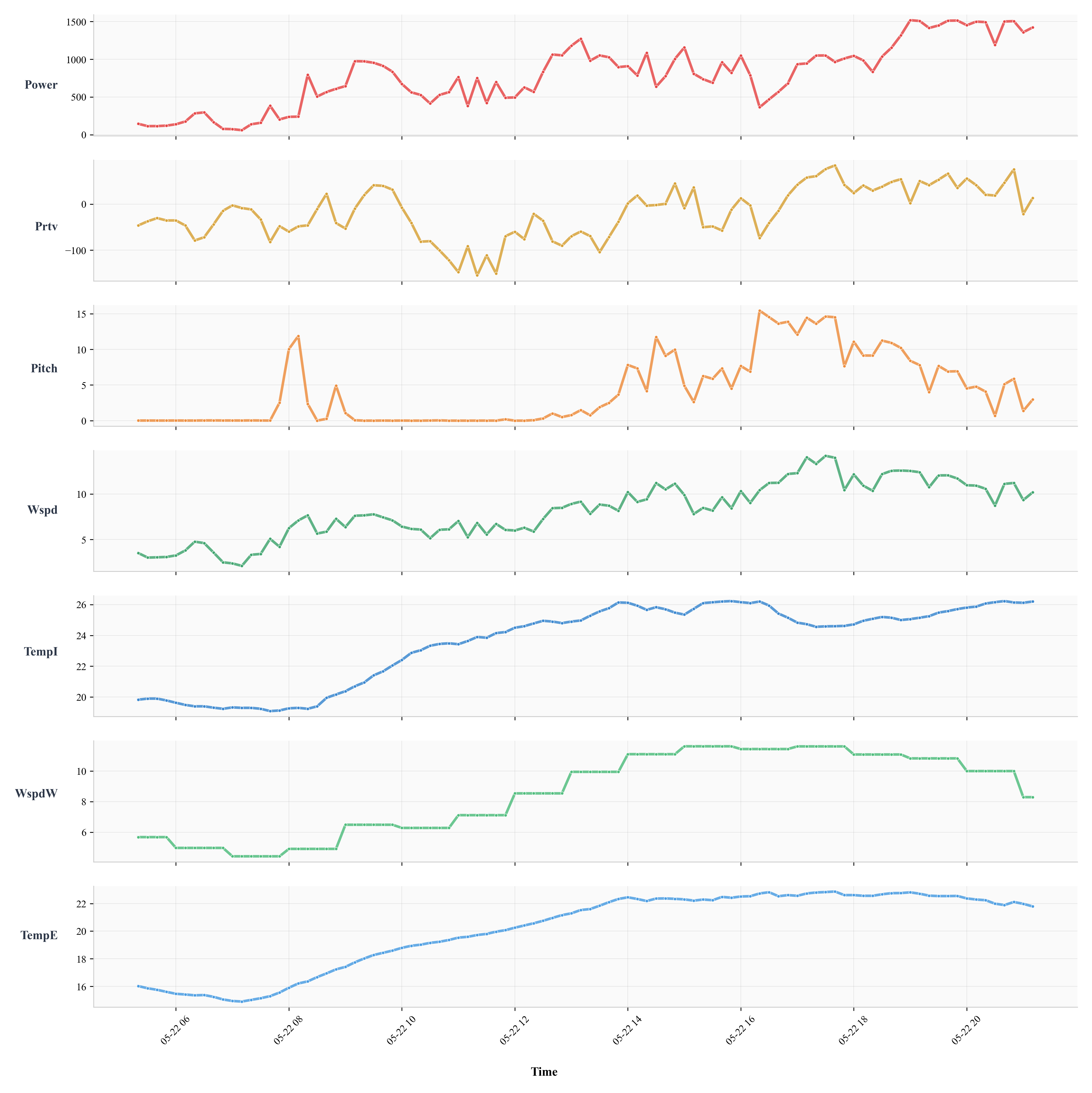} 
    \caption{The multivariate time series data sample.} 
    \label{fig:sample_timeseries} 
\end{figure}

To evaluate the effectiveness of our proposed LLM-driven feature selector, \textbf{Augur}, we conducted a comprehensive comparative analysis. We tested its performance against two baseline feature sets: one utilizing all available variables (All Features) and a univariate approach using only the most direct predictor (Wind Speed). These three feature sets were evaluated across four distinct forecasting architectures: MLP, LSTM, DLinear, and PatchTST. We report the Mean Squared Error (MSE) and Mean Absolute Error (MAE) for each experiment in Table~\ref{tab:forecasting_results_final}, with lower values indicating better performance.

The results clearly demonstrate the superiority of the Augur methodology for most models. By identifying a more informative and less noisy subset of variables, Augur consistently yielded the best performance for the MLP, LSTM, and DLinear architectures.while Augur provides a significant advantage for recurrent and linear models, advanced Transformer-based architectures like PatchTST may possess powerful internal mechanisms that are already highly effective at filtering and weighting information from a larger, unfiltered set of features.

In summary, the experiments validate that Augur serves as a powerful and effective feature selection framework. By identifying a causally-informed subset of variables, it consistently enhances the predictive accuracy of various conventional forecasting models, making a strong case for the integration of LLM-driven causal discovery into time-series analysis pipelines.

\subsubsection{LLM-as-a-Judge Evaluation}
\label{app:llm_judge_details}

To supplement our human evaluation and ensure a scalable, reproducible assessment of narrative quality, we implemented an "LLM-as-a-Judge" protocol. We employed a strong general-purpose model (e.g., \texttt{Gemini-2.5-Pro}) to act as an impartial evaluator. The judge was provided with the ground-truth time series data and the generated summaries from different models.

We instructed the judge to score each summary on a scale of 1 to 7 based on three specific dimensions. The specific prompt used is detailed below:

\vspace{0.5em}
\noindent\textbf{Role:} You are an expert data analyst and critic evaluating time series reports.

\noindent\textbf{Input:}
\begin{itemize}[leftmargin=*]
    \item \textbf{Time Series Data:} [Insert CSV snippet here]
    \item \textbf{Model Generated Summary:} [Insert Summary here]
\end{itemize}

\noindent\textbf{Evaluation Criteria:}
Please rate the summary on a scale of 1 (Poor) to 7 (Excellent) for each of the following metrics:
\begin{itemize}[leftmargin=*]
    \item \textbf{Ease of Understanding (EoU):} Is the text fluent, concise, and free of jargon?
    \item \textbf{Insightfulness (Ins.):} Does it identify meaningful patterns rather than just restating data points?
    \item \textbf{Causal Correctness (Corr.):} Do the attributed causes align with logical physical mechanisms (e.g., distinguishing cause from effect)?
\end{itemize}

\noindent\textbf{Output Format:} Return a JSON object with keys: \texttt{"EoU", "Ins", "Corr", "Reasoning"}.

\vspace{1em}
The quantitative results obtained from this evaluation protocol are summarized in Table~\ref{tab:llm_judge_results}.

\begin{table}[h]
\centering
\caption{LLM-as-a-Judge evaluation results.}
\label{tab:llm_judge_results}
\resizebox{0.48\textwidth}{!}{%
\begin{tabular}{@{}llcccc@{}}
\toprule
& & \multicolumn{4}{c}{\textbf{Evaluation Metrics}} \\
\cmidrule(lr){3-6}
\textbf{Dataset} & \textbf{Method} & \textbf{EoU} & \textbf{Ins.} & \textbf{Corr.} & \textbf{Avg.} \\
\midrule
\multirow{3}{*}{\textcolor{orange}{\textbf{Power}}}
 & GPT-4o & \textcolor{blue}{\underline{6.27}} & \textcolor{blue}{\underline{4.83}} & \textcolor{blue}{\underline{5.41}} & \textcolor{blue}{\underline{5.50}} \\
 & Qwen-3 & 6.18 & 4.56 & 4.72 & 5.15 \\
 & \textbf{Augur} & \textcolor{Magenta}{\textbf{6.34}} & \textcolor{Magenta}{\textbf{5.92}} & \textcolor{Magenta}{\textbf{6.18}} & \textcolor{Magenta}{\textbf{6.15}} \\
\midrule
\multirow{3}{*}{\textcolor{teal}{\textbf{Air}}}
 & GPT-4o & \textcolor{blue}{\underline{6.41}} & \textcolor{blue}{\underline{4.67}} & \textcolor{blue}{\underline{5.58}} & \textcolor{blue}{\underline{5.55}} \\
 & Qwen-3 & 6.29 & 4.39 & 4.85 & 5.18 \\
 & \textbf{Augur} & \textcolor{Magenta}{\textbf{6.53}} & \textcolor{Magenta}{\textbf{5.81}} & \textcolor{Magenta}{\textbf{6.07}} & \textcolor{Magenta}{\textbf{6.14}} \\
\bottomrule
\end{tabular}%
}
\end{table}

\subsection{Case Study: Wind Power Analysis}

\begin{table}[htbp]
  \centering
  \begin{threeparttable}
    \caption{Description of Variables}
    \label{tab:variables_correlation_matrix_simple}
    \small % Use smaller font for the table
    \begin{tabular}{l l}
      \toprule
      \textbf{Variable Name} & \textbf{Description} \\
      \midrule
      Power             & Active Power (actual generation output) \\
      Wind Speed        & Nacelle-measured wind speed \\
      Wind Dir          & Nacelle-measured wind direction \\
      Ext Temp          & External nacelle temperature \\
      Int Temp          & Internal nacelle temperature \\
      Nacelle Dir       & Nacelle direction (yaw angle) \\
      Pitch Rel         & Blade pitch relative value\\
      Atm Temp          & Atmospheric temperature at 2 meters \\
      Pressure          & Surface atmospheric pressure \\
      Humidity          & Relative humidity \\
      ERA5 Wind         & Reanalysis wind speed from ERA5 \\
      ERA5 Dir          & Reanalysis wind direction from ERA5 \\
      Precip Type       & Precipitation type (encoded)\\
      Pitch Angle       & Blade pitch angle\\
      \bottomrule
    \end{tabular}
    \end{threeparttable}
\end{table}

\paragraph{Initial graph construction}
We conduct a controlled comparison across correlation metrics (Pearson, Spearman, Kendall) and thresholds on per-sample windows of length $T{=}96$. For each sample, we  compute the correlation matrix over all numeric variables. We instantiate a candidate undirected graph by (i) linking $Patv$ to its top-$5$ variables ranked by $|\rho|$, (ii) adding pairwise edges among non-$Patv$ variables whenever $|\rho|\ge\tau$, and (iii) retaining only the connected component that contains $Patv$.

We report the average number of retained edges per sample as a proxy for search-space size. Edge counts decrease monotonically with $\tau$, and at any fixed $\tau$ the graphs induced by Pearson are densest, Kendall sparsest, with Spearman in between. Guided by this comparison, we fix \emph{Spearman} with $\tau{=}0.8$, yielding compact yet expressive candidate graphs for the subsequent LLM-guided causal inference stage.

\begin{table}[htbp]
\centering
\caption{Average connections by correlation threshold}
\begin{tabular}{lccccc}
\toprule
Method & 0.5 & 0.6 & 0.7 & 0.8 & 0.9 \\
\midrule
Spearman & 30.8 & 24.4 & 17.5 & 11.9 & 6.9 \\
Pearson  & 32.6 & 26.1 & 19.5 & 14.5 & 9.2 \\
Kendall  & 18.6 & 13.0 & 7.9  & 6.4  & 5.1 \\
\bottomrule
\end{tabular}
\end{table}

The analysis confirms that Active Power (\texttt{Patv}) is predominantly governed by two factors. It exhibits a very strong positive correlation with Wind Speed and a strong negative correlation with the blade \texttt{Pitch} angle. This relationship reflects the core physics of wind turbine operation: power output increases with wind speed until the pitch angle is adjusted to regulate the load. The ERA5 reanalysis wind speed shows a similar, though weaker, positive correlation.

Significant multicollinearity is evident among the environmental predictor variables. The various temperature readings (\texttt{Etmp}, \texttt{Itmp}, and \texttt{T2m}) are highly inter-correlated and share strong negative relationships with Surface Pressure (\texttt{Sp}). This indicates considerable redundancy among these atmospheric measurements.

These insights directly inform the modeling strategy. Wind Speed (\texttt{Wspd}) and Pitch Angle (\texttt{Pitch}) are confirmed as primary predictors for power forecasting. However, the redundancy observed among the temperature and pressure variables suggests that a careful selection or combination of these features is necessary to build a robust and efficient model.

\begin{figure}[h!]
\centering
\begin{tikzpicture}[
    scale=0.8,
    every node/.style={scale=0.85},
    >=latex,
    node distance=2cm and 2.4cm,
    var/.style={circle, draw=#1!80, fill=#1!20, minimum size=0.9cm, 
                font=\scriptsize\sffamily\bfseries, thick, drop shadow},
    source/.style={var=orange},
    intermediate/.style={var=teal},
    deep/.style={var=purple},
    target/.style={var=red},
    edge/.style={->, >=stealth, thick, draw=#1!70, line width=1pt},
    mainedge/.style={edge=gray},
    tempedge/.style={edge=teal},
    windedge/.style={edge=orange},
]
 
    \node[source] (sp) at (0, 0) {Sp};
    \node[source] (wspd) at (1.8, 0) {Wspd};
    \node[source] (wspd_w) at (3.6, 0) {Wspd\_w};
    \node[source] (wdir) at (5.4, 0) {Wdir};
    
    \node[intermediate] (etmp) at (1, -2.2) {Etmp};
    \node[intermediate] (t2m) at (2.7, -3) {T2m};
    \node[intermediate] (wdir_w) at (4.5, -2.6) {Wdir\_w};
    
    \node[deep] (itmp) at (2.2, -4.8) {Itmp};
    \node[deep] (relh) at (4.8, -5.2) {RelH};
    
    \node[target, minimum size=1.1cm] (patv) at (3.6, -7.2) {Patv};
    
    \begin{scope}[on background layer]
        \fill[orange!5, rounded corners=8pt] (-0.7, 0.6) rectangle (7.0, -0.6);
        \fill[teal!5, rounded corners=8pt] (0.3, -1.7) rectangle (5.2, -3.5);
        \fill[purple!5, rounded corners=8pt] (1.4, -4.3) rectangle (5.6, -5.7);
        \fill[red!5, rounded corners=8pt] (2.7, -6.7) rectangle (4.5, -7.7);
    \end{scope}
    
    \draw[tempedge] (sp) -- (etmp);
    \draw[mainedge, dashed] (sp) to[out=-45, in=135] (relh);
    
    \draw[windedge] (wspd) to[out=-90, in=150] (patv);
    \draw[windedge] (wspd_w) to[out=-90, in=120] (patv);
    \draw[windedge] (wdir) to[out=-90, in=60] (patv);
    \draw[windedge] (sp) to[out=-45, in=160] (patv);
    
    \draw[tempedge] (etmp) -- (t2m);
    \draw[tempedge] (etmp) to[out=-45, in=135] (itmp);
    \draw[tempedge] (etmp) to[out=-30, in=150] (wdir_w);
    \draw[tempedge] (etmp) to[out=-60, in=160] (relh);
    
    \draw[tempedge] (t2m) to[out=-45, in=90] (itmp);
    \draw[tempedge] (t2m) to[out=-60, in=120] (relh);
    
    \draw[mainedge] (itmp) to[out=-30, in=180] (relh);
    
    \draw[mainedge] (wdir_w) to[out=-90, in=45] (itmp);
    \draw[mainedge] (wdir_w) to[out=-45, in=90] (relh);
    
    \node[source, minimum size=0.5cm] at (6.8, -0.8) {};
    \node[font=\tiny\sffamily, right] at (7.1, -0.8) {\scriptsize Source};
    \node[intermediate, minimum size=0.5cm] at (6.8, -1.5) {};
    \node[font=\tiny\sffamily, right] at (7.1, -1.5) {\scriptsize Intermediate};
    \node[deep, minimum size=0.5cm] at (6.8, -2.2) {};
    \node[font=\tiny\sffamily, right] at (7.1, -2.2) {\scriptsize Deep};
    \node[target, minimum size=0.5cm] at (6.8, -2.9) {};
    \node[font=\tiny\sffamily, right] at (7.1, -2.9) {\scriptsize Target};
\end{tikzpicture}
\caption{Final causal DAG after cycle resolution}
\label{fig:final_dag}
\end{figure}
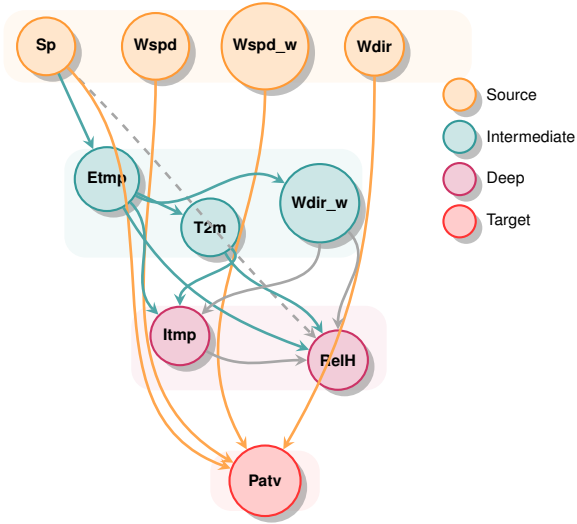

Figure \ref{fig:sample_timeseries} presents the multivariate time series data sample used in our case study . It plots the dynamics of key operational variables---including Active Power (\texttt{Power}), Wind Speed (\texttt{Wspd}), Pitch Angle (\texttt{Pitch}), and multiple temperature readings---over a single day. This observational data forms the empirical basis for the analysis in the following sections, where our model explains these dynamics using the causal rules defined by the discovered DAG (as shown in Figure \ref{fig:final_dag}).

\paragraph{Detected Cycles and Resolution Strategy}

The causal discovery agent identified multiple cycles in the wind power generation system. Below we present the first five cycles with their proposed resolutions.

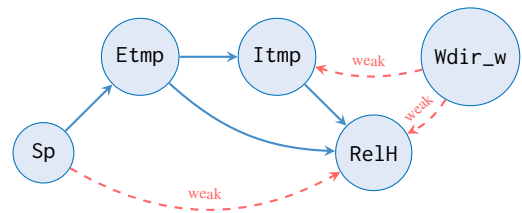
\begin{figure}[h!]
\centering
\begin{tikzpicture}[
    node distance=1.8cm,
    var/.style={circle,draw=RoyalBlue,fill=RoyalBlue!10,minimum size=0.8cm,font=\footnotesize\ttfamily},
    edge/.style={->,>=stealth,thick,RoyalBlue!70},
    weak/.style={->,>=stealth,thick,red!60,dashed},
    label/.style={font=\tiny,midway,above,sloped}
]
    % Nodes
    \node[var] (etmp) {Etmp};
    \node[var,right of=etmp] (itmp) {Itmp};
    \node[var,below right of=itmp] (relh) {RelH};
    \node[var,below left of=etmp] (sp) {Sp};
    \node[var,above right of=relh] (wdir) {Wdir\_w};
    
    % Core edges
    \draw[edge] (etmp) -- (itmp);
    \draw[edge] (etmp) to[bend right=20] (relh);
    \draw[edge] (itmp) -- (relh);
    \draw[edge] (sp) -- (etmp);
    
    % Weak edges (to be removed)
    \draw[weak] (sp) to[bend right=30] node[label] {weak} (relh);
    \draw[weak] (wdir) to[bend left=15] node[label] {weak} (relh);
    \draw[weak] (wdir) to[bend left=15] node[label] {weak} (itmp);
\end{tikzpicture}
\caption{Example of detected cycle with weak edges marked for removal}
\label{fig:cycle_example}
\end{figure}

\paragraph{Resolution Strategy}

\begin{table}[h!]
\centering
\caption{Edge Removal Frequency in Cycles}
\label{tab:edge_removal_freq}
\renewcommand{\arraystretch}{0.95}
\begin{tabular}{@{}lcccccc@{}}
\toprule
\textbf{Edge Type} & \multicolumn{6}{c}{\textbf{Removal Frequency}} \\
\cmidrule(lr){2-7}
& C1 & C2 & C3 & C4 & C5 & C6 \\
\midrule
Sp $\rightarrow$ RelH & \checkmark & \checkmark & -- & \checkmark & \checkmark & \checkmark \\
Wdir\_w $\rightarrow$ RelH & \checkmark & \checkmark & \checkmark & \checkmark & -- & \checkmark \\
Wdir\_w $\rightarrow$ Itmp & -- & -- & \checkmark & \checkmark & \checkmark & \checkmark \\
\bottomrule
\end{tabular}
\end{table}

\begin{table*}[ht!]
\centering
\caption{Cycle Resolution Summary}
\label{tab:cycle_resolution}
\footnotesize
\renewcommand{\arraystretch}{1.1}
\begin{tabular}{@{}p{0.5cm}p{4.5cm}p{2.5cm}p{7cm}@{}}
\toprule
\textbf{ID} & \textbf{Cycle Path} & \textbf{Edge to Remove} & \textbf{Justification} \\
\midrule
1 & Etmp$\to$Itmp$\to$RelH, Sp$\to$RelH, Sp$\to$Etmp & Sp$\to$RelH & Surface pressure influences humidity indirectly via temperature. Direct temperature-humidity links are mechanistically stronger. \\[3pt]

2 & Etmp$\to$Itmp$\to$RelH, Wdir\_w$\to$RelH, Wdir\_w$\to$Itmp, Sp$\to$RelH, Sp$\to$Etmp & Sp$\to$RelH & Sp's effect on RelH is mediated by temperature and wind conditions. Direct pathways from Etmp and Wdir\_w are more plausible. \\[3pt]

3 & Etmp$\to$Itmp$\to$RelH, Wdir\_w$\to$RelH, Wdir\_w$\to$Itmp, Etmp$\to$RelH, Sp$\to$RelH, Sp$\to$Etmp & Sp$\to$RelH & Sp primarily impacts humidity through temperature mediation. Presence of Sp$\to$Etmp supports this indirect pathway. \\[3pt]

4 & Etmp$\to$Itmp$\to$RelH, Wdir\_w$\to$RelH, Wdir\_w$\to$Itmp, Etmp$\to$RelH & Wdir\_w$\to$RelH & Wind direction's direct impact on humidity is less pronounced than temperature effects. Temperature links are stronger physical drivers. \\[3pt]

5 & Etmp$\to$Itmp$\to$RelH, Wdir\_w$\to$Itmp, Etmp$\to$RelH & Wdir\_w$\to$Itmp & External temperature (Etmp) is the primary driver of internal temperature, not wind direction. \\
\bottomrule
\end{tabular}
\end{table*}

\textbf{Primary criteria for edge removal:}
\begin{enumerate}[leftmargin=*]
\item \textbf{Physical plausibility:} Temperature gradients create stronger direct effects on humidity than pressure or wind direction.

\item \textbf{Causal mediation:} Indirect effects (e.g., Sp$\to$Etmp$\to$RelH) are preferred over spurious direct links.
\end{enumerate}
\textbf{Final DAG characteristics:} After removing identified weak edges, the resulting directed acyclic graph maintains wind-centric causal flow with no cycles, preserving mechanistically sound relationships essential for wind power prediction.
-
\paragraph{Causal Narrative Summary}

\textit{The data show a daytime increase in wind and temperature that drives large rises in Patv (power) and a concurrent warming (Itmp) with a midday dip in relative humidity.}

\bigskip 

\begin{description}

    \item[Finding 1: Power Generation Dynamics]
    \par\smallskip 
    \textbf{\textcolor{Maroon}{Pattern Observed:}} \\
    \textit{Patv rises sharply from early morning to afternoon/evening, tracking increases in Wspd and Wspd\_w (e.g., Patv: $\approx$144 $\to$ >1000; Wspd: $\approx$3.5 $\to$ 10--13).}
    
    \par\smallskip 
    \textbf{\textcolor{ForestGreen}{Causal Explanation (per DAG):}} \\
    Consistent with the DAG: rising \texttt{Wspd} directly increases \texttt{Patv} (\texttt{Wspd} $\to$ \texttt{Patv}) and increases \texttt{Wspd\_w} (\texttt{Wspd} $\to$ \texttt{Wspd\_w}), which in turn also increases \texttt{Patv} (\texttt{Wspd\_w} $\to$ \texttt{Patv}). The observed co-movement of \texttt{Pab} with \texttt{Wspd} is also expected by \texttt{Wspd} $\to$ \texttt{Pab}.

    \item[Finding 2: Internal Temperature Dynamics]
    \par\smallskip
    \textbf{\textcolor{Maroon}{Pattern Observed:}} \\
    \textit{Itmp (internal temperature) increases over the day ($\approx$19.8 $\to$ $\approx$26.2) following the rise in Patv and environmental temperatures.}
    
    \par\smallskip 
    \textbf{\textcolor{ForestGreen}{Causal Explanation (per DAG):}} \\
    Explained by DAG paths: higher \texttt{Etmp} raises \texttt{T2m} (\texttt{Etmp} $\to$ \texttt{T2m}) and \texttt{T2m} raises \texttt{Itmp} (\texttt{T2m} $\to$ \texttt{Itmp}), \texttt{Etmp} also increases \texttt{Patv} (\texttt{Etmp} $\to$ \texttt{Patv}) and \texttt{Patv} raises \texttt{Itmp} (\texttt{Patv} $\to$ \texttt{Itmp}), and rising \texttt{Wspd} increases \texttt{Wspd\_w} which also raises \texttt{Itmp} (\texttt{Wspd} $\to$ \texttt{Wspd\_w} $\to$ \texttt{Itmp}). These combined causal routes account for the daytime warming of \texttt{Itmp}.

    \item[Finding 3: Relative Humidity Dynamics]
    \par\smallskip
    \textbf{\textcolor{Maroon}{Pattern Observed:}} \\
    \textit{Relative humidity falls through midday ($\approx$0.22 $\to$ $\approx$0.18) while temperatures rise, then partially recovers later.}
    
    \par\smallskip
    \textbf{\textcolor{ForestGreen}{Causal Explanation (per DAG):}} \\
    Per the DAG, \texttt{Etmp} directly affects \texttt{RelH} (\texttt{Etmp} $\to$ \texttt{RelH}); additionally \texttt{Itmp} influences \texttt{RelH} (\texttt{Itmp} $\to$ \texttt{RelH}) and \texttt{Wspd}-driven \texttt{Wspd\_w} also affects \texttt{RelH} (\texttt{Wspd} $\to$ \texttt{Wspd\_w} $\to$ \texttt{RelH}). Thus the midday RH drop is attributable to higher \texttt{Etmp} and \texttt{Itmp} (and concurrent \texttt{Wspd\_w} changes) via the DAG-prescribed links.

\end{description}

\begin{figure*}[htbp]
  \centering
  \includegraphics[width=\textwidth]{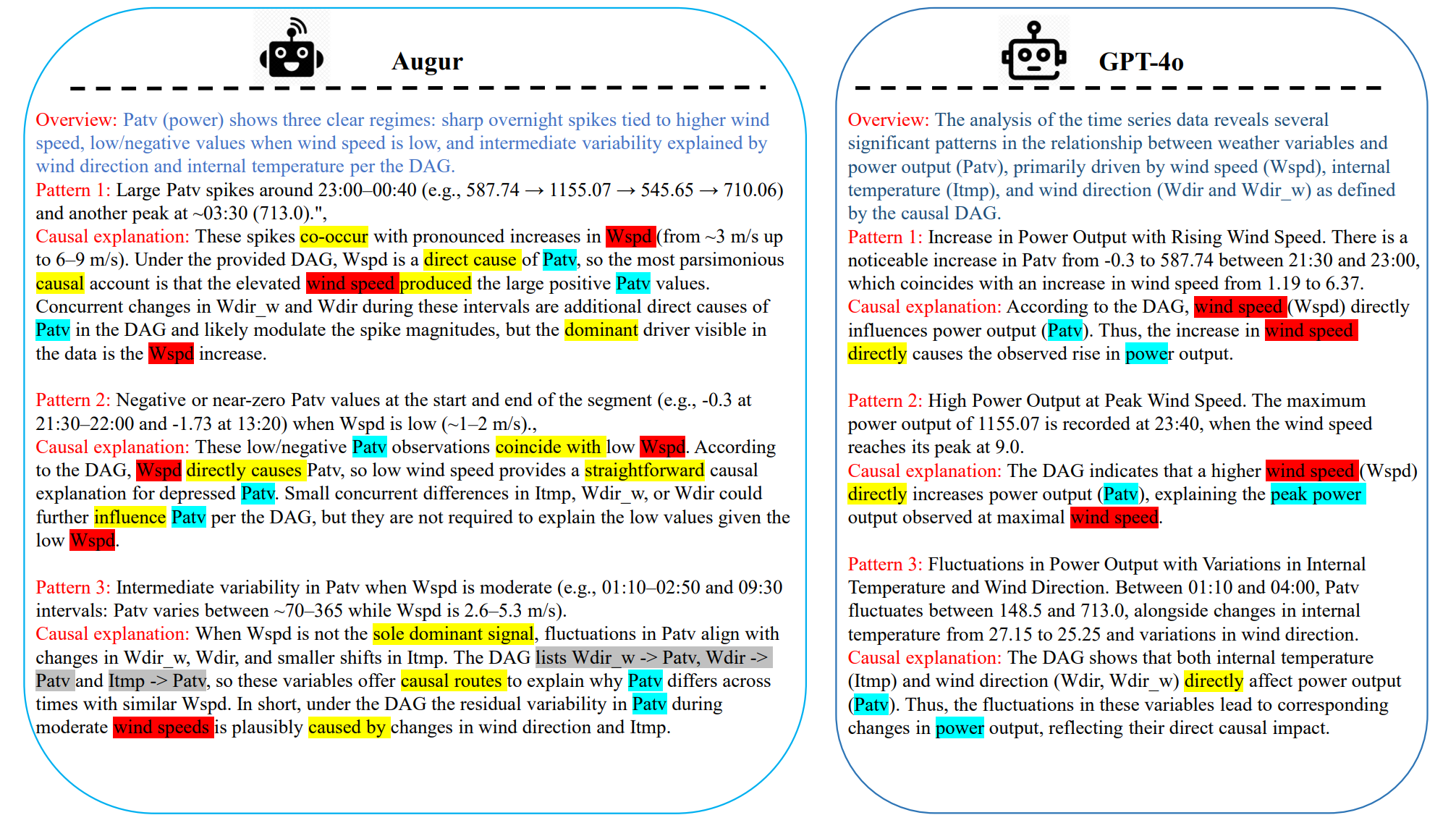}
  \caption{Side-by-side comparison of causal narrative outputs for the same  segment.}
  \label{fig:cs_vs_gpt4o}
\end{figure*}
This generated causal summary provides critical, actionable insights for predictive modeling. By validating the system's true causal drivers (such as \texttt{Wspd} and \texttt{TempE}) and their pathways, the analysis confirms the variables belonging to the theoretical Causal Markov Blanket. This allows for the construction of a sparse, robust, and generalizable feature set while safely excluding redundant proxy variables.

Most critically, the analysis moves beyond simple correlation to prevent modeling errors. For instance, by identifying that \texttt{Power} causes internal temperature (via the path \texttt{Power} → \texttt{TempI}), the summary explicitly instructs us to \textbf{exclude} \texttt{TempI} as a predictor for \texttt{Power}. A standard model based on correlation alone would likely misuse this variable, learning a spurious relationship that degrades predictive stability.

Finally, this summary reveals the complex, multi-path dynamics required for truly robust forecasting. The insight that \texttt{TempI} is driven by \textit{both} environmental heat (\texttt{Etmp} → \texttt{T2m} → \texttt{TempI}) and operational heat (\texttt{Power} → \texttt{TempI}) allows a model to correctly anticipate system states---such as a high internal temperature on a cold but windy day---that a purely correlative model would fail to predict. This causal grounding directly translates to a more physically accurate and reliable forecasting system.

\section{Extensibility}

As shown in Figure~\ref{fig:cs_vs_gpt4o}, Augur produces DAG-grounded explanations that identify wind speed as the dominant causal driver of power and provide causal routes for residual variability, whereas GPT-4o tends to restate correlations without consistently grounding claims in the graph.

\paragraph{Beyond Causality.} A key advantage of our framework is the inherent modularity and extensibility of the generated narrative. The textual format of our causal explanation allows for seamless integration with other forms of time-series analysis, creating a richer auxiliary modality for any downstream task. For instance, the narrative can be programmatically augmented with other structured insights, such as automatically inserting pre-computed \textbf{statistical properties} like the Spearman correlation coefficients. Similarly, results from classical time-series decomposition can be integrated to explicitly state \textbf{temporal dynamics} like periodicities or trends. By concatenating these diverse textual representations, we can construct a holistic, multi-faceted summary that captures not only the "why" (causality) but also the "what" (statistics) and the "how" (patterns). This enriched textual modality provides a far more complete contextual understanding for any language-model-based downstream system, positioning our framework as a central component in a broader, hybrid time-series analysis ecosystem.

\paragraph{Societal Impact.}
The deployment of Augur introduces a paradigm shift towards transparent and interpretable time series forecasting, which is particularly critical for high-stakes domains. In sectors such as \textbf{healthcare}, providing explicit causal rationales for patient vitals monitoring can assist clinicians in making informed, life-saving decisions rather than relying on opaque "black-box" alerts. Similarly, in \textbf{energy grids} and \textbf{financial markets}, understanding the causal drivers behind volatility allows for more robust risk management and policy formulation. 

\paragraph{Efficiency and Controllability.} While our framework leverages a powerful teacher model for the initial, one-time generation of the training corpus, we deliberately employ a smaller, supervised fine-tuned (SFT) student model for all downstream tasks. This strategic choice is driven by critical considerations of efficiency, economy, and controllability, which are paramount in real-world time-series applications. Direct, repeated inference with a frontier model like a hypothetical GPT-5 would be prohibitively expensive and slow for the high-throughput processing often required in time-series analysis. 

By distilling the teacher's complex reasoning capabilities into a specialized student agent, we achieve a system that is not only orders of magnitude more cost-effective and faster at inference, but also more controllable. The SFT approach allows us to create a deterministic, self-contained artifact that can be deployed reliably in production environments without reliance on external APIs, ensuring stable performance and predictable behavior. This distillation process, therefore, represents a pragmatic yet powerful method to harness the reasoning power of state-of-the-art LLMs while meeting the practical constraints of operational time-series analysis.

\subsection{Prompt Example}
\textbf{Prompt 1} generates pairwise causal hypotheses for correlated variables. These hypotheses are then passed to \textbf{Prompt 2}, which assembles them into a global structure and resolves cycles to form a valid Directed Acyclic Graph (DAG). Finally, \textbf{Prompt 3} uses this validated DAG to synthesize a grounded narrative that explains key patterns observed in the time series data.

\clearpage
\begin{tcolorbox}[
    skin=enhanced,                   
    width=\textwidth,               
    arc=4mm,                          
    colback=white,                   
    colframe=NavyBlue,              
    colbacktitle=NavyBlue,           
    coltitle=white,                  
    fonttitle=\bfseries,             
    drop shadow={NavyBlue!50!Black},  
    title= \faLightbulb \space Prompt 1: Pairwise Causal Hypothesis Generation 
]

\textbf{\textcolor{RoyalBlue}{ROLE:}} \\
You are an expert in \textit{[Your Domain, e.g., "financial markets"]} and a specialist in causal inference.

\smallskip 
\textbf{\textcolor{TealBlue}{CONTEXT:}} \\
I am analyzing data from a \textit{[System or Process Name]} to build a Causal Directed Acyclic Graph (DAG). I have identified a significant statistical correlation between two variables and need to determine their causal associations.

\smallskip 
\textbf{\textcolor{OliveGreen}{VARIABLE DEFINITIONS:}}
\begin{description}
    \item[Variable A:] \textit{[Variable A Name] - [Clear, concise definition...]}
    \item[Variable B:] \textit{[Variable B Name] - [Clear, concise definition...]}
\end{description}

\smallskip 
\textbf{\textcolor{Plum}{INPUT DATA:}} \\
Correlation between \textit{[Var A]} and \textit{[Var B]}: \textit{[e.g., "Spearman's rho = +0.85"]}

\smallskip 
\textbf{\textcolor{Maroon}{TASK:}} \\
Evaluate the following causal hypotheses based on first principles...

\smallskip 
\textbf{\textcolor{ForestGreen}{HYPOTHESES:}}
\begin{itemize}
    \item \texttt{A -> B:} ...
    \item \texttt{B -> A:} ...
    \item \texttt{Confounder:} ...
    \item \texttt{Correlation Only:} ...
\end{itemize}

\smallskip 
\textbf{\textcolor{MidnightBlue}{OUTPUT FORMAT:}} \\
Provide a JSON object with keys: "reasoning" and "conclusion".

\end{tcolorbox}

\clearpage
\bigskip 

\begin{tcolorbox}[
    skin=enhanced,                   
    width=\textwidth,
    arc=4mm,
    colback=white,
    colframe=NavyBlue,
    colbacktitle=NavyBlue,
    coltitle=white,
    fonttitle=\bfseries,
    drop shadow={NavyBlue!50!Black},
    title= \faSitemap \space Prompt 2: Global Graph Assembly \& Cycle Resolution 
]

\textbf{\textcolor{RoyalBlue}{ROLE:}} \\
You are an expert in systems modeling and graph theory, specializing in the validation of causal structures.

\smallskip
\textbf{\textcolor{TealBlue}{CONTEXT:}} \\
I have performed pairwise causal analysis to generate a set of directed edges representing a system's hypothesized causal structure in the domain of \textit{[Your Domain]}. I need you to validate this structure.

\smallskip
\textbf{\textcolor{Plum}{INPUT: LIST OF DIRECTED EDGES}} \\
\textit{[Paste all inferred directed edges from Stage 1 here, one per line.]}

\begin{verbatim}
VarA -> VarB
VarC -> VarA
VarB -> VarC
...
\end{verbatim}

\smallskip
\textbf{\textcolor{Maroon}{TASK:}}

\begin{enumerate}
    \item \textbf{Identify Cycles:} Analyze the provided edges and explicitly identify any cycles.
    \item \textbf{Propose Resolution:} For each cycle, propose which single edge is the "weakest link" and should be removed.
    \item \textbf{Justify Proposal:} Provide a clear, logical justification for your choice.
\end{enumerate}

\smallskip
\textbf{\textcolor{MidnightBlue}{OUTPUT FORMAT:}} \\
Provide a structured response listing identified cycles and your justified recommendations. If no cycles exist, state that "The graph is a valid DAG."

\end{tcolorbox}

\clearpage
\bigskip 

\begin{tcolorbox}[
    skin=enhanced,                    
    width=\textwidth,
    arc=4mm,
    colback=white,
    colframe=NavyBlue,
    colbacktitle=NavyBlue,
    coltitle=white,
    fonttitle=\bfseries,
    drop shadow={NavyBlue!50!Black},
    title= \faChartLine \space Prompt 3: Causal Analysis \& Summary from Time Series Data 
]

\textbf{\textcolor{Maroon}{TASK:}} \\ 
Your task is to analyze the provided multivariate time series data to identify the 2-3 most significant patterns or events. Then, write a concise narrative summary that explains your findings using the causal associations defined in the Causal DAG.

\smallskip
\textbf{\textcolor{Plum}{INPUTS:}} 
\par\smallskip 
\textbf{1. Causal DAG:} \textit{(This graph is the "rule book" for causation...)}
\begin{verbatim}
[Paste your DAG here, one edge per line, e.g.:]
Wspd -> Patv
Patv -> Itmp
Etmp -> Itmp
\end{verbatim}
\par\smallskip
\textbf{2. Core Variable Time Series:} \textit{(Provide a downsampled or key segment...)}
\begin{verbatim}
[Paste your time series data here, for example:]
Timestamp, Wspd, Patv, Itmp
2025-09-12 12:00, 8.1, 1.2, 45.1
2025-09-12 12:05, 15.2, 2.5, 45.5
...
\end{verbatim}

\smallskip
\textbf{\textcolor{Maroon}{INSTRUCTIONS:}} 
\begin{enumerate}
    \item \textbf{Analyze First:} Examine the raw time series to find the most important patterns...
    \item \textbf{Explain with DAG:} For each significant pattern you identify, construct a causal explanation...
    \item \textbf{Causal Fidelity is Crucial:} You must not infer any cause-and-effect relationship...
\end{enumerate}

\smallskip
\textbf{\textcolor{MidnightBlue}{OUTPUT:}} \\ 
Produce a concise summary. Start with a one-sentence overview, followed by bullet points. Each bullet point should first \textbf{describe a key pattern} you found in the data and then \textbf{explain its cause(s)} based on the DAG.

\end{tcolorbox}

\end{document}